\documentclass[letterpaper,twocolumn,10pt]{article}
\usepackage{usenix2019_v3}
\AtBeginDocument{%
  \providecommand\BibTeX{{%
    \normalfont B\kern-0.5em{\scshape i\kern-0.25em b}\kern-0.8em\TeX}}}

\usepackage{kotex}

\usepackage{amsmath}
\usepackage[ruled,lined,linesnumbered]{algorithm2e}
\usepackage{amssymb}
\usepackage{mathtools}
\usepackage{xcolor}

\usepackage{graphicx}
\usepackage{float}
\usepackage{booktabs} 
\usepackage{comment}
\usepackage{setspace}
\usepackage{amsmath,amssymb,array,threeparttable}
\usepackage{multirow}\usepackage{xcolor,colortbl}
\usepackage{hyperref}
\usepackage{graphicx}
\usepackage{float}
\usepackage{url}
\usepackage{graphicx}
\usepackage{endnotes}
\usepackage{breakurl}
\usepackage{enumitem} 
\usepackage{tikz}
\usepackage{epstopdf}
\usepackage{multirow}
\usepackage{flushend}
\usepackage[export]{adjustbox}
\usepackage{menukeys}
\usepackage{verbatim}
\usepackage{balance}
\usepackage{bbm}
\usepackage{xcolor}
\usepackage{subcaption}
\usepackage{siunitx}
\usepackage{booktabs}       %
\usepackage{caption}        
\usepackage{array}    
\usepackage{tabularx}  

\definecolor{darkgreen}{RGB}{0,100,0}
\definecolor{mediumgreen}{RGB}{0,128,0}
\definecolor{lightgreen}{RGB}{144,238,144}
\definecolor{limegreen}{RGB}{50,205,50}
\definecolor{forestgreen}{RGB}{34,139,34}

\newcolumntype{P}[1]{>{\centering\arraybackslash}p{#1}}
\newcolumntype{M}[1]{>{\centering\arraybackslash}m{#1}}

\DeclareMathOperator*{\argmin}{arg\,min}
\newcommand{\gs}[1]{\textcolor{black}{#1}}

\newcommand{\JY}{\textcolor{black}}
\newcommand{\TODO}{\textcolor{black}}
\newcommand{\REV}{\textcolor{black}}

\usepackage{titlesec}
\titleformat{\section}
  {\normalfont\large\bfseries}{\thesection}{1em}{}
\titleformat{\subsection}
  {\normalfont\normalsize\bfseries}{\thesubsection}{1em}{}

\newcommand{\sysname}[1]{\textbf{\sf HAFS}}

\newenvironment{packeditemize}{
\begin{list}{$\bullet$}{
\setlength{\itemsep}{0pt}
\addtolength{\labelwidth}{10pt}
\setlength{\leftmargin}{12pt}
\setlength{\listparindent}{\parindent}
\setlength{\parsep}{2pt}
\setlength{\topsep}{0pt}}}
{\end{list}
}

\usepackage{titlesec}
\titleformat{\subsubsection}
  {\normalfont\normalsize\itshape}{\thesubsubsection}{1em}{}
\titlespacing*{\subsubsection}{0pt}{3.25ex plus 1ex minus .2ex}{0ex plus .2ex}

\newcommand{\bcircled}[2][black]{%
  \tikz[baseline=(char.base)]{%
    \node[shape=circle, fill=#1, text=white, draw=none,
          inner sep=1pt,             
          minimum size=0.8em,        
          font=\sffamily\small\bfseries] (char) {#2};}}

\AtBeginDocument{%
  \setlength{\abovedisplayskip}{2.2pt plus 1pt minus 0pt}%
  \setlength{\belowdisplayskip}{2.2pt plus 1pt minus 0pt}%
  \setlength{\abovedisplayshortskip}{1pt plus 1pt minus 0pt}%
  \setlength{\belowdisplayshortskip}{1pt plus 1pt minus 0pt}%
}

\begin{document}

\title{\large Coordinated Networking for On-Device Agent-augmented Real-Time Communication}

\author{
{\rm Goodsol Lee}\textsuperscript{1,2}\footnotemark[1] \quad
{\rm Juheon Yi}\textsuperscript{1}\footnotemark[2] \quad
{\rm Jinglu Wang}\textsuperscript{1} \quad
{\rm Haowen Xu}\textsuperscript{1} \quad
{\rm Saewoong Bahk}\textsuperscript{2} \quad
{\rm Yan Lu}\textsuperscript{1}\\[0.4em]
\textsuperscript{1}Microsoft Research Asia \quad
\textsuperscript{2}Seoul National University\\
}

\maketitle
\vspace{-15pt}
\begin{abstract}
AI agents are enabling a new paradigm of agent-augmented real-time communication (RTC), where humans focus on high-level collaboration, while agents autonomously retrieve, analyze, and generate information in real time to support their interactions.
As existing cloud-based agents suffer from privacy risks and unscalable server costs, \textit{on-device agent-augmented RTC} offers a promising alternative.
However, this on-device paradigm introduces a new  networking challenge: contention between concurrent traffic flows generated by humans (for live video streaming) and agents (for sending context files for analysis). 
We design \sysname{}, a framework to ensure both high live video quality and low agent response latency in agent-augmented RTC apps. 
We achieve the goal through an \emph{app-guided multi-flow transport} approach, where a unified app-layer orchestrator jointly controls the sending rates of live video and agent context flows based on their heterogeneous app requirements.
Our prototype built atop WebRTC and llama.cpp demonstrates that \sysname{} outperforms baselines, achieving 1.5$\times$ higher video quality while reducing agent response time by 31\%.
\end{abstract}

\renewcommand{\thefootnote}{\fnsymbol{footnote}}
\footnotetext[1]{Work done during Goodsol's internship at Microsoft Research Asia.}
\footnotetext[2]{Corresponding author.}
\renewcommand{\thefootnote}{\arabic{footnote}}

\section{Introduction} 
\label{sec:1-intro}

\begin{figure}[t]
    \centering
    \begin{center}          \includegraphics[width=1\columnwidth]{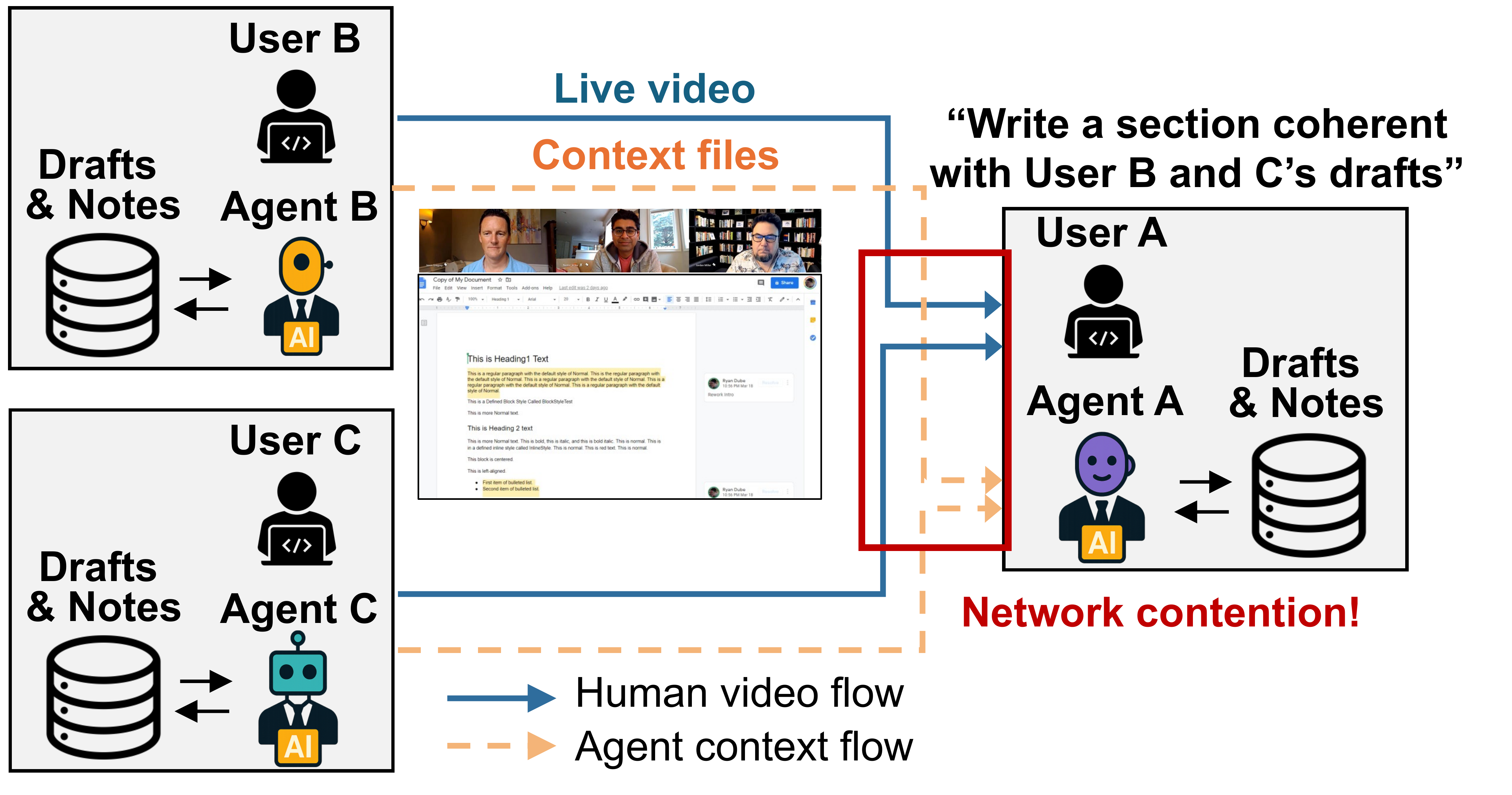}
        \vspace{-15pt}
        \caption{
        \JY{On-device agent-augmented RTC scenario and its network contention problem.}
        }
        \label{fig:1-scenario}
    \vspace{-20pt}
    \end{center}
\end{figure}

With the emergence of AI agents,\footnote{An AI agent is a Large Language Model (LLM)-driven system that can retrieve/analyze multi-modal inputs, reason over them, and execute actions autonomously to accomplish user requests~\cite{kim2025cost}.} real-time communication applications (RTC apps) are evolving into \emph{agent-augmented RTC}.
Assume a group of corporate employees are having a remote video call to collaboratively review a 1,000-page legal contract they have drafted in parts.
As the team cross-checks existing clauses or adds new ones, the agent automatically identifies and analyzes relevant contexts, and suggests a coherent example draft (Figure~\ref{fig:1-scenario}).
This allows employees to avoid the overhead of manually inspecting each other's work, significantly boosting their productivity (see detailed scenarios in \S\ref{subsec:2-scenarios}).
We envision this paradigm enabling useful services across various domains such as collaborative coding, media co-creation, and remote healthcare.

Commercial RTC platforms already deploy early forms of such agents.
For example, Microsoft Teams' Facilitator summarizes conversations and suggests follow-up questions~\cite{teams_facilitator}.
However, these systems largely depend on cloud-based agents, raising two practical deployment issues.
(i) \emph{Privacy}. Users may feel reluctant to upload proprietary or sensitive data (e.g., business documents, personal chat logs) to the cloud.
(ii) \emph{Cost}. 
As the number of users and concurrent queries grow, cloud-based inference incurs unscalable GPU costs.
A promising alternative is \emph{on-device} agent-augmented RTC, where each agent runs locally on the user’s device and communicates with peers to exchange necessary contexts to process queries.
This allows the platform provider to reduce cloud GPU costs and also keep sensitive user data encrypted during network transmission.
This becomes increasingly feasible with recent Small Language Models (e.g., Llama3.1-8B~\cite{dubey2024llama}, Qwen2.5-7B~\cite{bai2025qwen2}) that can run on-device while achieving comparable accuracy to cloud-scale LLMs through task-specific fine-tuning~\cite{roziere2023code}.

Despite its potential, realizing on-device agent-augmented RTC is highly challenging.
The core challenge lies in coordinating concurrent, heterogeneous network flows generated by humans and agents.
During the RTC session, humans stream their live video (\emph{human video flows}), while agents concurrently transmit multiple context segments in response to peer queries (\emph{agent context flows}).
\JY{
In particular, it is preferable for the sender agent to prefill the raw context into a Key-Value (KV) cache offline and stream it to the receiver to eliminate heavy LLM prefill from agent response time (e.g., reducing the latency from 105 to 16 seconds, \S\ref{subsec:3-base-pipeline}).
However, this heavily inflates traffic volume: for example, an 8k-token raw context is merely a few KBs, but its KV cache expands to $\approx$80~MB even after compression.
As these flows contend over the shared network bandwidth, careful coordination is essential to guarantee high video quality and low agent response latency.
}
Moreover, such agent flows are spawned continuously as users issue queries throughout the collaboration process.
Compared to traditional RTC sessions where users occasionally exchange a single file~(e.g., PowerPoint slides) or small-sized chat messages, this multi-stream, continuous flow workload is significantly more challenging from a networking perspective.

We present \sysname{} (\underbar{H}uman-\underbar{A}gent \underbar{F}low \underbar{S}cheduling), a coordinated networking framework for on-device agent-augmented RTC apps.
\sysname{} efficiently coordinates heterogeneous flows to guarantee both high live video streaming quality and low agent response latency.  
\sysname{} is designed for practical deployment across a wide range of devices, networks, and RTC frameworks (specifically, the de facto standard WebRTC) and multi-user video conferencing architectures with only app-side modifications.

Designing \sysname{} requires solving two key challenges:
\begin{packeditemize}
    \item \textbf{Human-agent flow contention} (\S\ref{subsubsec:3-human-agent-flow-contention}). 
    In commercial RTC platforms (e.g., Microsoft Teams, Zoom), human video flows use delay-based congestion control~\cite{carlucci2016analysis, salsify_nsdi18, lee2021demystifying} to minimize streaming delay, while agent context flows use loss-based congestion control~\cite{ha2008cubic, grieco2004performance} to maximize throughput at the cost of high queuing delay. 
    \JY{
    Without coordination, the agent context flow aggressively fills the network buffer, causing the human video flow to suffer from queuing delays. 
    As a result, the video congestion controller significantly reduces its bitrate (e.g., by 53\% in Zoom), and it recovers conservatively once throttled (e.g., taking $>$30\,s to fully restore).
    }
    
    \item \textbf{Multi-agent flow contention} (\S\ref{subsubsec:3-multi-agent-flow-contention}). 
    When multiple contexts need to be streamed to process a single query, \JY{data channels of conventional RTC platforms} schedule them in an app-agnostic, round-robin manner.
    As a result, flows suffer from inter-stream contention and inflated queuing delays. 
    This leaves the receiver’s compute resources underutilized while waiting for delayed flow completions, which significantly increases the agent response time (e.g., 56\% longer than the optimal baseline).
\end{packeditemize}    

\JY{
\sysname{} addresses these challenges via an \emph{app-guided multi-flow transport} (\S\ref{subsec:4-approach}).
Specifically, a unified app-layer orchestrator maintains global visibility over all outgoing video and agent flows and jointly controls their sending rates based on app requirements, thereby simultaneously achieving high video quality and low agent response time.
Conventionally, end-host flow scheduling struggles to achieve precise rate control due to delayed feedback over long RTT links.
Modern RTC architectures, however, present a unique opportunity: as users are connected through intermediate relay servers (Selective Forwarding Units) whose locations are determined based on their geographic distributions~\cite{sfu, sfu2}, the end-to-end path is effectively split into sender–SFU and SFU–receiver links with significantly shorter RTTs (e.g., p95 latency under 50~ms~\cite{cloudflare_calls_anycast_webrtc_2024}).
Thus, our approach enables precise per-flow rate control akin to in-network scheduling (e.g., router-level priority-based flow scheduling in datacenters~\cite{al2010hedera,liu2025pyrrha}), while also remaining highly practical to deploy by requiring modifications at the end-host's app layer.
}

We realize our approach with two key components:
\begin{packeditemize}
    \item \textbf{Human-agent flow coordinator} (\S\ref{sec:5-coordinator-human-agent}) 
    \JY{
    controls the video and agent flow sending rates to simultaneously maximize video quality and agent throughput.
    To maximize link utilization without degrading video quality, we design a novel multi-flow rate control mechanism which leverages the video frame-level queuing as its guiding signal.
    Specifically, it ramps up and sustains the aggregate rate until the frame-level queuing delay approaches—without exceeding—the app-specified frame deadline (e.g., 150~ms~\cite{meng2022achieving, dhawaskar2023converge}).
    Given this determined total rate, it strictly provisions the required rate for the live video and allocates the remaining slack capacity to agent flows. 
    }
    
    \item \textbf{Multi-agent flow coordinator} (\S\ref{sec:6-coordinator-multi-agent}) 
    \JY{
    schedules the transmission order of concurrent agent flows within the agent flow-allocated rate to minimize the agent response time.
    We formulate this as a two-machine flow shop problem~\cite{garey1976complexity} with joint consideration of KV cache sizes, LLM decoding latencies, and bandwidths of sender-SFU links. 
    As LLM response length (which dictates decoding latency) remains unknown before decoding, we predict it using a lightweight single-step LLM decoding, a method proven feasible for maximizing batching in datacenter LLM serving~\cite{jin2023s, zheng2023response}.
    }
\end{packeditemize}

\JY{
We prototyped \sysname{} atop WebRTC~\cite{website:WebRTC} and llama.cpp~\cite{llamacpp},
and evaluated it in diverse real-world environments spanning various mobile/edge devices (MacBook Pro, Samsung Galaxy S25, NVIDIA Jetson), networks (Wi-Fi~6, 5G), and LLMs (from Qwen~\cite{bai2025qwen2} and Llama~\cite{dubey2024llama} families). 
\gs{\sysname{} significantly outperforms the state-of-the-art baselines, achieving \TODO{1.5$\times$} higher video quality (VMAF~\cite{li2018vmaf} score comparable to that of video-only streaming) while reducing agent response time by 31\%.}}

\section{On-Device Agent-augmented RTC}
\label{sec:2-background}


\subsection{Application Scenarios}
\label{subsec:2-scenarios}

\noindent {\bf Corporate document co-authoring.}
A group of corporate employees is tasked with writing a 1,000-page legal contract.
After dividing the work and drafting sections individually, they meet through a remote video call for collaborative review.
During the review of each member's draft, team members verify in parallel whether the clauses are comprehensively articulated and add necessary clauses while maintaining coherence.
Given the document's massive volume, manually cross-referencing all details is burdensome; an agent assists by analyzing relevant sections, and providing responses.

\vspace{3pt}
\noindent {\bf Memory-augmented meeting.}
A group of graduate students attends a week-long academic conference to present a paper. 
During the event, each student's AR glasses continuously record their conversations with other attendees.
In a subsequent remote lab-wide meeting, as each student shares their experience, the advisor and peers simultaneously ask multiple questions, such as whether industry representatives mentioned recruiting opportunities or requesting a summary of feedback received on the paper.
Since the student cannot remember all conversations in detail, an agent retrieves the stored conversation logs, analyzes them, and provides responses.


\subsection{Why On-Device Agent?}
\label{subsec:2-on-device}

\REV{
Commercial RTC platforms are supporting collaborative tasks that require shared context (e.g., coding and document editing)~\cite{ms_liveshare, google_docs}, and are introducing agents to assist collaboration in real-time~\cite{teams_facilitator}.
While they mostly run agents on cloud servers, we envision on-device agents for two key reasons.
}
(i) \emph{Privacy}. Unlike cloud agents that require users to expose sensitive data (e.g., personal conversation logs and proprietary documents) in a raw format to the RTC platform provider's servers, on-device agents keep user data encrypted throughout the entire transmission path, aligning with recent end-to-end encryption practices in RTC platforms~\cite{zoom_e2ee_meetings, teams_e2ee_meetings, webex_e2ee_meetings_calling, googlemeet_encryption}.
(ii) \emph{Cost}. By offloading agent inference to user devices, RTC providers can circumvent cloud GPU costs, which typically scale linearly with the number of session participants.
Recent small language models~(SLMs), such as Qwen2.5-3B~\cite{bai2025qwen2} and Llama3.1-8B~\cite{dubey2024llama}, make on-device inference more feasible; they achieve both high throughput and accuracy, especially when fine-tuned to specific domains~\cite{belcak2025small, roziere2023code}.

\subsection{Workload Characterization}
\label{subsec:2-workload}

\begin{figure}[t]
    \centering
\includegraphics[width=1\columnwidth]{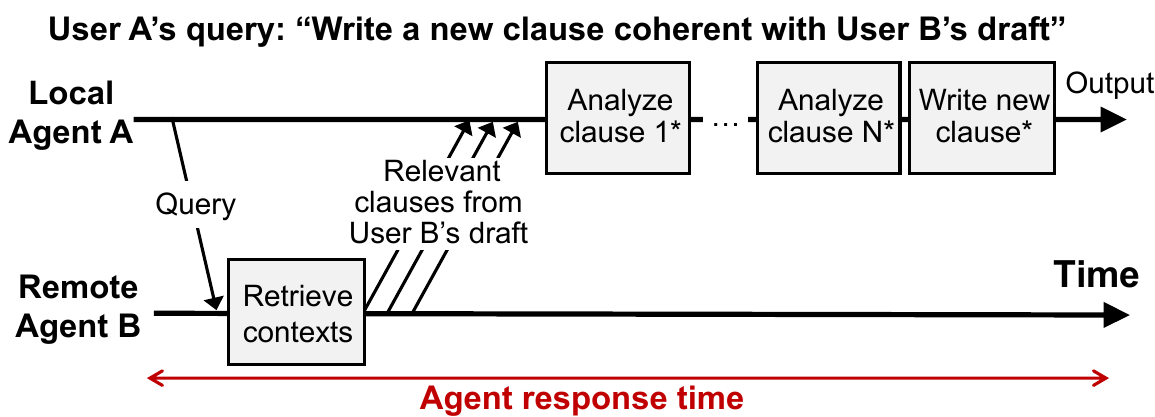}
    \vspace{-15pt}
    \caption{
    Agent workflow for corporate document co-authoring scenario. * denotes LLM inference.
    }
    \label{fig:2-agent-workflow}
    \vspace{-15pt}
\end{figure}

As shown in Figure~\ref{fig:1-scenario}, on-device agent-augmented RTC apps involve two distinct types of concurrent network flows:

\vspace{3pt}
\noindent\textbf{Human video flows} 
for human-to-human communication, requiring high-quality live video (e.g., $>$10~Mbps bitrate and $<$150~ms frame delay~\cite{meng2022achieving, dhawaskar2023converge}) for seamless interaction.

\vspace{3pt}
\noindent \textbf{Agent context flows}
for agent-to-agent communication to retrieve and analyze relevant contexts to answer user queries.
Figure~\ref{fig:2-agent-workflow} shows an example of an agent workflow to process a user's query in the corporate document co-authoring scenario.
The local agent first forwards the query to peer agents to request relevant contexts.
\REV{
Upon receiving each context, the agent runs LLM inference to generate partial responses and aggregates them as output response.\footnote{\REV{We assume this two-stage processing pipeline for two reasons:
(i) on-device LLMs have limited context windows, making it difficult to process long contexts in a single pass; and
(ii) waiting for all contexts before decoding precludes network--compute pipelining and increases agent response time.
We discuss context granularity optimization in \S\ref{sec:9-discussion}.
Aggregation uses an additional LLM call to produce a user-friendly summary of per-context analysis outputs, incurring negligible latency compared to per-context decoding.}}
}
Agent response time is defined as the delay between the submission of the user query and the agent's final output response.
\REV{
Latency requirements vary with task type and complexity~\cite{abbas2022understanding,dabrowski201140} and should be adjusted to the target use case.
For example, in our empirical study, users found the agent responsive when it answered simple factual verification over short contexts (e.g., a 1k-token dialogue) within $\approx$10s, and complex long-context summarization (e.g., condensing an 8k-token report into a 300-token paragraph) within $\approx$30s.
}
\section{Motivation}
\label{sec:3-motivation}

In this section, we introduce the base system pipeline design for agent-augmented RTC (\S\ref{subsec:3-base-pipeline}), and analyze the network contention problems that emerge between the human video and agent context flows within this architecture (\S\ref{subsec:3-contention}).

\vspace{-10pt}
\subsection{Base System Pipeline Design}
\label{subsec:3-base-pipeline}

Building the base pipeline for an on-device agent-augmented RTC system requires navigating a key design space: how to split LLM inference.
Specifically, we must determine how to divide the workload between the sender and receiver agents across two dimensions: (i) which agent performs the context prefill (processing input context tokens and storing them as a key-value (KV) cache) and at what stage, and (ii) which agent performs decoding (i.e., using the KV cache to generate the output response in an autoregressive manner).
In the following, we explore this design space to optimize agent response time and overall system resource efficiency.

\subsubsection{LLM Inference Latency Breakdown}
\label{subsubsec:3-inference-latency-breakdown}

\begin{figure}[t]
\centering
    \begin{minipage}[t]{0.23\textwidth}
        \centering
\includegraphics[width=\textwidth]{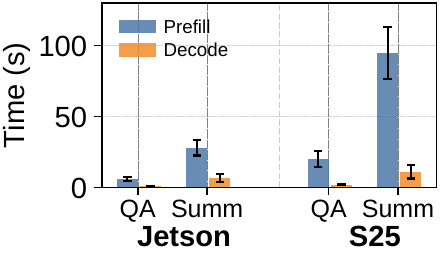}
        \vspace{-20pt}
        \captionof{figure}{LLM inference latency breakdown.}
        \label{fig:3-prefill-decode}
    \end{minipage}
    \hfill
    \begin{minipage}[t]{0.23\textwidth}
        \centering
        \includegraphics[width=\textwidth]{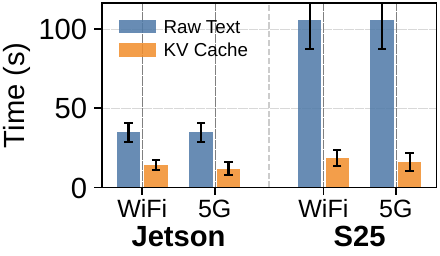}
        \vspace{-20pt}
        \captionof{figure}{Latency of raw text vs. KV cache streaming.}
        \label{fig:3-raw-kvzip}
    \end{minipage}
\vspace{-10pt}
\end{figure}

We analyzed the on-device LLM inference latency to guide our inference splitting decision.
Figure~\ref{fig:3-prefill-decode} shows the LLM prefill and decode latencies of Qwen2.5-3B~\cite{bai2025qwen2} model for document question answering (QA) and summarization tasks.
These were evaluated using the MuSiQue~\cite{trivedi2022musique} and QMSum~\cite{zhong2021qmsum} datasets on NVIDIA Jetson Orin Nano and Samsung Galaxy S25 (details in \S\ref{subsec:8-setup}).
We find that the prefill stage is the primary bottleneck: for summarization task, it accounts for 25 out of 35 seconds (71\%) of the total response time on Jetson, and 67 out of 73 seconds (92\%) on S25.
This is because edge devices have limited compute parallelism compared to server GPUs (e.g., Jetson Orin: 1,024 vs. H100: 14,592 CUDA cores~\cite{NVIDIA2025JetsonOrinnx, nvidia_h100_whitepaper}). 
Consequently, they struggle to parallelize the massive matrix multiplications of the prefill phase~\cite{song2024powerinfer}, making it the dominant bottleneck.

\vspace{-10pt}
\subsubsection{How to Split LLM Inference?}
\label{subsubsec:3-llm-inference-split}

Our LLM latency analysis shows that to optimize response time and resource efficiency, the sender agent should prefill the contexts offline, stream relevant KV caches upon query, and leave decoding to the receiver agent.
This design is particularly well-suited for on-device agent-augmented RTC due to two key app characteristics.

\begin{packeditemize}
\item 
In many cases, context generation and query issuance occur at different times (e.g., conversations are logged when a student attends a conference, and queries are later issued by peers in a meeting).
Thus, it is possible to prefill the context in advance, eliminating a substantial portion of the LLM inference latency at runtime.

\item 
\REV{
We primarily target collaboration scenarios where multiple users query a shared context concurrently (e.g., peers reviewing a colleague's draft or asking about his past conversations).
In such scenarios, prefilling the context once at the sender and letting receivers that need it fetch the KV cache and decode in parallel is preferable in terms of system-wide latency and energy consumption~\cite{nvidia_dynamo_lmcache_docs_2026, vllm_prefix_design_2026}, as network transmission consumes significantly less energy than GPU processing (evaluated in \S\ref{subsec:8-eval-deep-dive}).
}
\end{packeditemize}

Potential drawbacks of this design include storage and network transmission overheads for large KV caches.
For instance, the raw KV cache of Qwen2.5-3B yields 36 KB per token, 
resulting in 288 MB for an 8k-token context.
However, they are not prohibitive for the following reasons.

\begin{packeditemize}
\item 
Query-agnostic KV cache compression efficiently reduces size without sacrificing accuracy; for example, KVZip~\cite{kim2025kvzip} reduces the KV cache size by 70\% for Qwen2.5-3B/7B and Llama3.1-8B, maintaining ≥95\% and ≥98\% accuracy for QA and summarization (Appendix~\ref{appendix:kvzip-accuracy}).

\item 
Modern network bandwidth makes KV cache streaming much faster than local prefill.
For example, Wi-Fi~6 and 5G easily support >200–300 Mbps per-user throughput~\cite{yang2022mobile, k2024unveiling}. 
As validated in our testbed (Figure~\ref{fig:3-raw-kvzip}), streaming the KV cache over a 258 ± 19 Mbps Wi-Fi~6 cuts agent response time by up to 56\% on Jetson and 84\% on S25. We observe similar trends for NPUs (evaluation in \S\ref{subsec:8-eval-deep-dive}).
\end{packeditemize}

\subsection{Network Contention Problem}
\label{subsec:3-contention}

While our base pipeline is promising for reducing agent response time, our measurements on commercial RTC platforms reveal critical network contention issues.
Even with sufficient bandwidth, the concurrent transmission of multiple flows induces two distinct types of contention:
(i)~\textit{human-agent flow contention}~(\S\ref{subsubsec:3-human-agent-flow-contention}), where KV cache streaming degrades live video quality, and
(ii)~\textit{multi-agent flow contention}~(\S\ref{subsubsec:3-multi-agent-flow-contention}), where multiple KV caches streamed concurrently delay each other and lead to low compute resource utilization.

\vspace{-10pt}
\subsubsection{Background on RTC Frameworks}
\label{subsubsec:3-implementation}

Our base system pipeline can be implemented on top of WebRTC (de facto standard RTC platform) as follows. 
Commercial RTC apps also share a similar architecture~\cite{lee2021demystifying, michel2022enabling}.
           
\vspace{3pt}                      
\noindent
\textbf{RTP for human video flows.}
Live video is delivered via RTP (Real-Time Transport Protocol)~\cite{wikipedia_rtp} over UDP. 
For interactive communication (e.g., $<$150~ms frame delay~\cite{meng2022achieving}), the receiver sends periodic RTCP feedback reporting per-packet delay and loss.
Based on this, delay-sensitive Google Congestion Control~(GCC)~\cite{carlucci2016analysis} dynamically adapts the video bitrate to reduce  self-inflicted queuing. 

\vspace{3pt}
\noindent
\textbf{SCTP for agent context flows.}
KV caches are transmitted through SCTP~(Stream Control Transmission Protocol)~\cite{stewart2002sctp} over UDP, which provides reliable delivery with loss-based congestion control~\cite{grieco2004performance}.
It aggressively grows the congestion window to maximize throughput, at the cost of filling network buffers.
By default, the SCTP stack operates independently and remains uncoordinated with the RTP stack, as data traffic in conventional RTC apps is typically intermittent and small in size (e.g., chat messages or one-time file transfers).

\vspace{3pt}
\noindent
\textbf{SFU-based multi-party architecture.} 
In a multi-party RTC session, each user’s human and agent context flows are streamed to a central SFU (Selective Forwarding Unit) server~\cite{michel2025scalable} which relays them to peers, thus eliminating redundant peer to peer transmissions.

\subsubsection{Human-Agent Flow Contention}
\label{subsubsec:3-human-agent-flow-contention}

\begin{figure}[t]
\centering
    \begin{subfigure}[t]{0.23\textwidth}
        \centering
        \includegraphics[width=\textwidth]{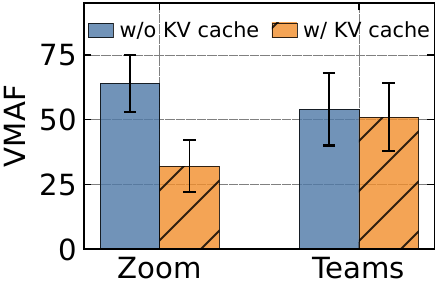}
        \caption{Video quality with and without KV cache streaming.}
        \label{fig:3-commercial-vmaf}
    \end{subfigure}
    \hfill
    \begin{subfigure}[t]{0.23\textwidth}
        \centering
        \includegraphics[width=\textwidth]{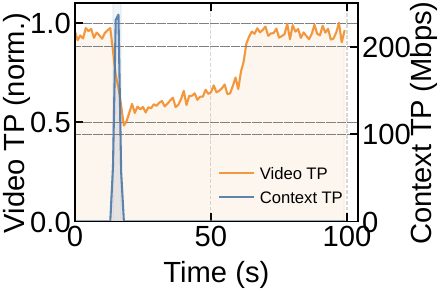}
        \caption{Slow video bitrate recovery after congestion.}
        \label{fig:3-zoom_timeline}
    \end{subfigure}
	\vspace{-8pt}
    \caption{Human-agent flow contention.} 
\vspace{-15pt}
\label{fig:3-commercial}
\end{figure}

We first evaluate whether commercial RTC platforms (Zoom~\cite{zoom}, Microsoft Teams~\cite{microsoft_teams}, Google Meet~\cite{google_meet}) can support concurrent video streaming and KV cache transmission.
We send an 80~MB KV cache (equivalent to an 8k-token context for Qwen2.5-3B, compressed with KVZip~\cite{kim2025kvzip}) alongside live video via built-in file sharing
under real-world Wi-Fi~6 and 5G networks (details in~\S\ref{subsec:8-setup}). 
We captured the network traffic via Wireshark to analyze the underlying transmission patterns.
We find that none of the three platforms adequately handles this workload:

\begin{figure}[t]
\centering
    \begin{minipage}[t]{0.23\textwidth}
        \centering
        \includegraphics[width=\textwidth]{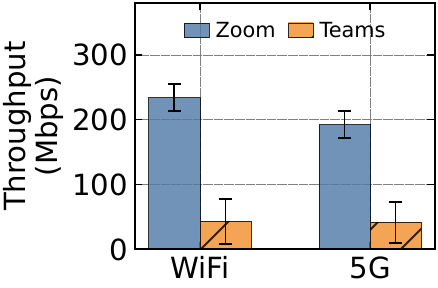}
        \vspace{-20pt}
        \caption{Agent context flow in Zoom and Teams.}
        \label{fig:3-commercial-throughput}
    \end{minipage}
    \hfill
    \begin{minipage}[t]{0.23\textwidth}
        \centering
        \includegraphics[width=\textwidth]{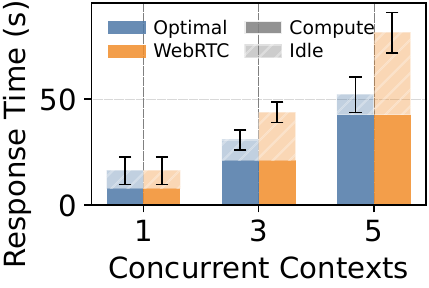}
        \vspace{-20pt}
        \captionof{figure}{Multi-agent flow contention in WebRTC.}
    \label{fig:3-webrtc_measurements}
    \end{minipage}
\vspace{-17pt}
\end{figure}

\begin{packeditemize}
    \item \noindent{\bf Zoom} transmits a KV cache by aggressively filling the network buffer,
    without any coordination with the live video stream.
    Consequently, the transport congestion controller of the live video stream experiences increased packet delays due to queuing, and reduces congestion window size even when the available network bandwidth is sufficient.
    This leads to significant degradation in video quality (e.g., 53\% VMAF~\cite{li2018vmaf} score drop as shown in Figure~\ref{fig:3-commercial-vmaf}).
    Moreover, as delay-based congestion controllers recover the rate conservatively to prevent oscillations,
    the video quality remains degraded for a long period even after the agent flow terminates (e.g., $>$30 seconds as shown in Figure~\ref{fig:3-zoom_timeline}).

    \item \noindent{\bf Teams} minimizes live video quality degradation by throttling the agent flow transmission (e.g., transmitting only $\approx$34~MB every 10 seconds).
    However, this results in severely reduced network bandwidth utilization and agent context flow throughput; for instance, only 17\% of Zoom’s throughput under the same Wi-Fi environment (Figure~\ref{fig:3-commercial-throughput}).

    \item \noindent{\bf Meet} does not allow direct file transfers to preserve video quality, and supports only sharing Google Drive links.
\end{packeditemize}

\subsubsection{Multi-Agent Flow Contention}
\label{subsubsec:3-multi-agent-flow-contention}

Commercial RTC platforms support only one data flow at a time alongside live video.
\REV{
To analyze contention across multiple agent flows, we streamed multiple KV caches (60-115 MB and response lengths 17-260 tokens) from a desktop server to an S25 over WebRTC SCTP data channel on 260~Mbps Wi-Fi 6 network.
The receiver begins decoding only after a context's full KV cache arrives and processes KV caches in arrival order, since batching and preemption are infeasible on edge GPUs with limited memory capacity and bandwidth~\cite{yi2020heimdall}.
}
Figure~\ref{fig:3-webrtc_measurements} shows that agent response latency increases significantly with more concurrent agent flows. 
This is mainly due to SCTP’s app-agnostic flow scheduling. 
As it is unaware of the agent flows' sizes and decoding latencies (proportional to the response length), it multiplexes them in a round-robin manner at transport packet granularity (typically 1.5 KB~\cite{choi2023large}).
Consequently, all flows suffer from delayed completion times, causing compute resources to remain idle for long periods.
For example, with five concurrent flows, the agent response time is 56\% longer compared to an optimal (81 vs. 52s).

\section{\sysname{} Design}
\label{sec:4-design}

\subsection{Approach}
\label{subsec:4-approach}

We aim to simultaneously achieve high video quality (high bitrate and low frame delay) and low agent response time. 
The key to achieving this lies in efficiently orchestrating multiple concurrent flows with heterogeneous app requirements.

\vspace{3pt}
\noindent
\textbf{Limitations of existing approaches.}
In-network solutions, such as router-level priority-based flow scheduling in datacenters~\cite{al2010hedera,liu2025pyrrha}, can precisely enforce per-flow rates at the bottleneck link.
However, they are impractical for RTC apps where traffic traverses independently managed network segments that necessitate cross-vendor coordination~(e.g., laptops, Wi-Fi APs, and ISP routers).
End-host congestion control algorithms (CCAs) bypass this deployment hurdle, but existing approaches mostly optimize a single flow in isolation~\cite{arun2018copa, salsify_nsdi18, wang2024pudica}.
While a few recent works target multi-flow coordination in RTC, they remain largely app-agnostic; for example, Microsoft Teams simply throttles data traffic~(\S\ref{subsubsec:3-human-agent-flow-contention}), and FSE~\cite{islam2022real} splits rate across RTP and SCTP pursuing transport-level fairness.
Such coarse-grained coordination sufficed in traditional RTC scenarios, as data traffic was typically small, intermittent, and delay-tolerant~(e.g., chat messages or file transfers), allowing systems to focus exclusively on preserving video quality.
However, in agent-augmented RTC, agent flows are bulky, frequent, and latency-sensitive, necessitating fine-grained coordination to simultaneously achieve high video quality and low agent response times. 

\vspace{3pt}
\noindent
\textbf{Our approach.} To address this challenge, we design an \textit{app-guided multi-flow transport}. 
At its core is a unified app-layer orchestrator that maintains global visibility over all outgoing human video and agent context flows from an RTC app.
Leveraging this visibility, it controls the total sending rate to keep video frame-level queuing within the app-specified deadline. 
It then strictly provisions the required rate for live video, and distributes the remaining slack capacity among agent flows to maximize network-compute pipelining and minimize agent response time.
Conventionally, end-host solutions struggle to achieve flow control as precise as that of in-network approaches, as long RTTs between distant endpoints and delayed feedback drop bandwidth estimation accuracy.
However, modern RTC architectures naturally mitigate this limitation, as users are connected through an intermediate SFU that is dynamically deployed based on their geographic distribution~\cite{sfu, sfu2}.
Thus, the end-to-end path is split into sender–SFU and SFU–receiver links with significantly shorter RTTs (e.g., p95 latency under 50~ms~\cite{cloudflare_calls_anycast_webrtc_2024}), enabling effective flow scheduling even at the app layer.

\subsection{Design Challenges}
\label{subsec:4-challenges}

\noindent
\textbf{C1: How to maximize agent throughput without interfering with live video?}
Ideally, agent flows should consume the residual bandwidth (i.e., the available bandwidth excluding the required video rate) to maximize link utilization and avoid interference with live video.
However, strictly matching this agent flow rate is inherently challenging. 
Maximizing link utilization requires aggressively inflating queues at the cost of packet delay (as in conventional TCP CCAs~\cite{cardwell2016bbr, ha2008cubic}), which inevitably increases the queuing delay experienced by live video. 
Conversely, being overly sensitive to packet delay causes overreactions to minor jitter~\cite{arun2018copa, wang2024pudica} and bandwidth underutilization~\cite{arun2022starvation}.

\vspace{3pt}
\noindent
\textbf{C2: How to maximize video quality under self-induced agent flow contention?}
Conventional delay-sensitive RTP CCAs cannot distinguish queuing caused by genuine bandwidth drops from self-induced contention by co-existing agent flows. 
This confusion causes unnecessary video rate drops and slow recovery, even when there is enough bandwidth (\S\ref{subsubsec:3-human-agent-flow-contention}). 
Therefore, \sysname{} needs a way to accurately distinguish these congestion types and enhance video bitrates.

\vspace{3pt}
\noindent
\textbf{C3: How to schedule multiple agent flows under unknown compute workload?} 
As observed in \S\ref{subsubsec:3-multi-agent-flow-contention}, multi-agent flow scheduling must be app-aware, accounting for both the size and decoding latency (LLM's output response length) of individual agent context flows.
This is challenging, however, as the response length is inherently unknown prior to decoding.

\subsection{Key Design Ideas}
\label{subsec:4-ideas}

\noindent
\textbf{I1: Frame-level queuing-aware agent rate control.} 
We control the agent rate based on the video frame-level queuing delay.
We ramp up the agent rate until the frame-level queuing delay closely approaches, but does not exceed, the app-specified deadline.
This allows us to build up sufficient queues to maximize link utilization while minimizing interference with the live video.
For this, \sysname{} estimates frame-level queuing from existing transport feedback signals, instead of adding customized feedback that interferes with RTP CCAs.

\vspace{3pt}
\noindent \textbf{I2: Agent-probed bandwidth for video rate control.}
As agent context flows are typically transmitted at rates well above the maximum video bitrate, their received rate serves as an available bandwidth capacity estimate, which is unavailable to conventional delay-based RTP CCAs.
A high received rate indicates ample bandwidth availability, allowing the CCA to ignore delay increases, prevent false video rate drops, and accelerate recovery.
Coupled with the agent flow's frame-level queuing-aware rate control, this approach simultaneously maximizes link utilization and video quality.

\vspace{3pt}
\noindent \textbf{I3: Lightweight response length prediction from prebuilt KV cache.}
Upon receiving a user query, the sender agents perform a lightweight single-step LLM decoding using the corresponding prebuilt KV caches to predict the expected response length-an approach proven feasible by recent studies~\cite{jin2023s, zheng2023response} aiming to maximize batching in datacenter LLM serving.
Although these estimates inherently contain prediction errors, multi-agent flow scheduling remains robust because its optimality depends solely on relative ordering of decoding latencies among flows, rather than absolute values.

\begin{figure}[t]
    \centering
        \centering
\includegraphics[width=1\columnwidth]{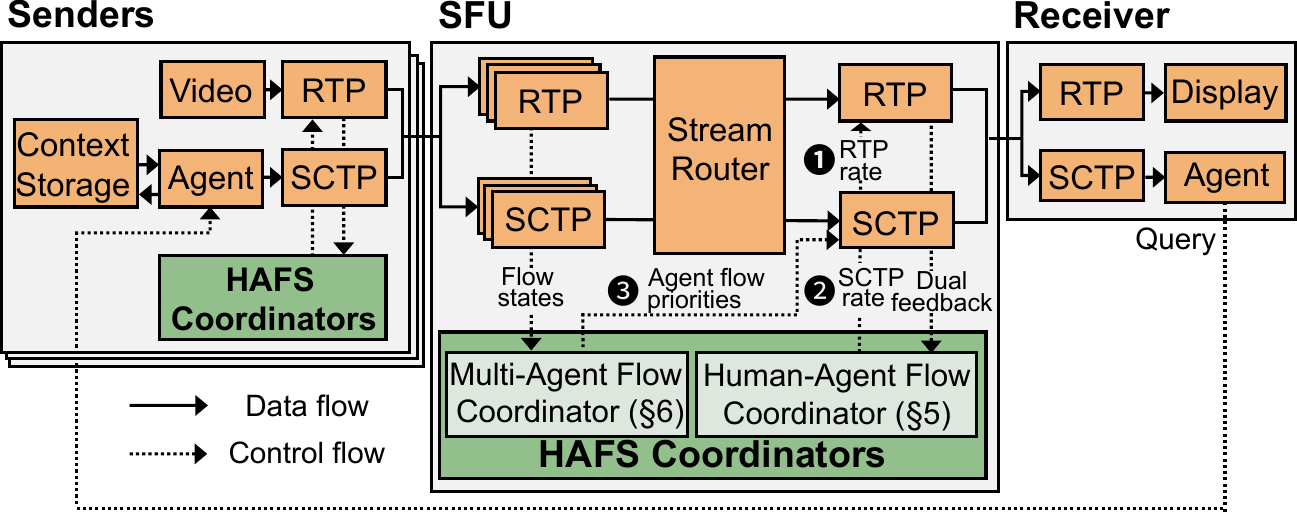}
            \vspace{-20pt}
            \caption{\JY{\sysname{} system architecture (orange components: data plane, green components: control plane).}}
            \label{fig:4-architecture}
            \vspace{-10pt}
\end{figure}

\subsection{System Architecture}
\label{sec:4-overview}

Figure~\ref{fig:4-architecture} shows the overall architecture of \sysname{}.

\vspace{3pt}
\noindent
\textbf{Data plane.}
The SFU relays each sender's live video stream over RTP, selecting the forwarded quality to match the receiver's available bandwidth~\cite{lin2022gso}. When the receiver issues a query, its agent forwards the query to each sender's agent, which retrieves the relevant pre-computed KV caches from its local context storage via a retrieval technique (e.g., similarity-based retrieval~\cite{lewis2020retrieval, guu2020realm}) and transmits them to the SFU over SCTP. The SFU aggregates the KV caches from all senders and forwards them to the receiver for LLM decoding.

\vspace{3pt}
\noindent
\textbf{Control plane.} 
\REV{
When a receiver-side query requires KV cache transfer from the sender, two coordinators operate in both the sender and the SFU to orchestrate concurrent human video and agent context flows.
}
(i) \emph{Human--agent flow coordinator}~(\S\ref{sec:5-coordinator-human-agent}) 
\JY{leverages RTP and SCTP feedback to measure video frame-level queuing and controls the total sending rate to prevent video quality degradation. 
Within this rate budget, it guarantees the RTP bandwidth (\bcircled{1}) and allocates any residual capacity to SCTP (\bcircled{2}).}
(ii) \emph{Multi-agent flow coordinator}~(\S\ref{sec:6-coordinator-multi-agent}) 
determines the transmission order across multiple agent context flows which maximizes network-compute pipelining, by jointly considering the flow sizes, predicted LLM decoding latencies, and sender-SFU link bandwidths~(\bcircled{3}).

\section{Human-Agent Flow Coordinator}
\label{sec:5-coordinator-human-agent}

\subsection{Overview}

\noindent
\textbf{Goal and challenges.} 
Human-agent flow coordinator jointly controls the RTP and SCTP sending rates to simultaneously maximize video quality and agent throughput.
Our approach is to design a joint rate control mechanism guided by the frame-level queuing, as introduced in \S\ref{subsec:4-ideas}.
Realizing this, however, raises two technical challenges.
First, accurately estimating video frame-level queuing (i.e., the queuing delay accumulated within each frame interval under current sending rate) is non-trivial, as the RTP video frame transmission typically spans only a small fraction of the frame interval (e.g., 3 out of 33 ms).
In the presence of cross-traffic or wireless channel fluctuations, frame-level queuing measured over such short period becomes a poor predictor over the full frame interval. 
\REV{Pudica~\cite{wang2024pudica} recently addressed this issue by dynamically pacing the video frames based on estimated bandwidth utilization, but this interferes with the native pacing of underlying RTP CCAs~\cite{huang2025ace, webrtc_probe_controller}.}
Second, even with accurate frame queuing measurements, determining how to efficiently control the sending rates is non-trivial.

\vspace{3pt}
\noindent \textbf{Our solution.} 
\emph{Dual feedback-based frame queuing estimator} (\S\ref{subsec:5-dual}) accurately measures frame-level queuing by combining RTP feedback packets with SCTP's per-packet ACKs, which densely populate the idle interval after RTP frame transmissions.
Guided by the measurement, \emph{frame queuing-guided multi-flow rate controller} (\S\ref{subsec:5-app-guided}) employs an MI/AI-MD mechanism to control the total sending rate.
It rapidly pushes the queuing delay closely toward (but without violating) the app-specified frame deadline, thus ensuring maximum link utilization without degrading video quality.
Within the determined aggregate rate, it strictly provisions the required rate for live video, and allocates the slack capacity to SCTP.

\subsection{Dual Feedback-based Frame Queuing Estimator}
\label{subsec:5-dual}

\begin{figure}[t]
\centering
    \includegraphics[width=1\columnwidth]{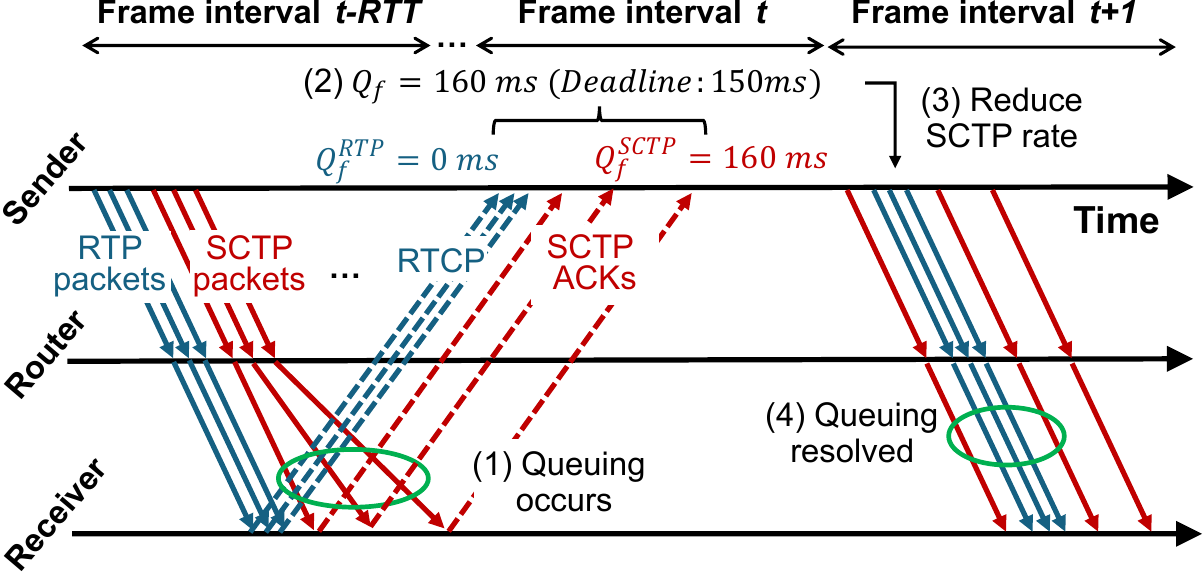}
    \vspace{-20pt}
    \caption{Human-agent flow coordinator workflow.}
\label{fig:5-human-agent-workflow}
\vspace{-13pt}
\end{figure}

\noindent \textbf{RTP-based frame-level queuing measurement.} Within a frame interval (length $L$)~(e.g., 33~ms at 30~fps), queuing during the transmission of video frame $f$ is computed as
\begin{equation}
\label{eq:5-Q-rtp}
    Q_f^{\mathrm{RTP}} = D_f - D_{\min},
\end{equation}
where $D_f$ is the one-way transmission delay of $f$ (measured at the receiver and reported back to the sender), and $D_{\min}$ is the minimum one-way packet delay observed within a recent window (which approximates the propagation delay without queuing).
Note that $D_f-D_{\min}$ cancels out the sender-receiver clock offset in one-way delay measurements.

\vspace{3pt}
\noindent \textbf{SCTP-based frame-level queuing measurement.}
After the transmission of frame $f$, we measure the frame-level queuing for the remainder of the frame interval from SCTP ACKs.
To compute $Q_f^{\mathrm{SCTP}}$, we need two additional considerations.
First, SCTP ACKs do not include one-way delay measurements as opposed to RTP.
Second, we must explicitly exclude any queuing caused by the preceding RTP frame, since this has already been accounted for in Equation~(\ref{eq:5-Q-rtp}). 
Taking these into account, we measure $Q_f^{\mathrm{SCTP}}$ as

\vspace{-5pt}
\begin{equation}
\label{eq:5-Q-sctp}
    Q_f^{\mathrm{SCTP}} = \max\!\left(0,\; \overline{\mathrm{RTT}}_f - \mathrm{RTT}_f^{\mathrm{ref}}\right),
\end{equation}
where $\overline{\mathrm{RTT}}_f$ is the average RTT over all SCTP ACKs in the frame interval after RTP frame transmission, and $\mathrm{RTT}_f^{\mathrm{ref}}$ is the reference RTT measured at the end of RTP frame $f$, which captures the queue occupancy created by the RTP packets.
By subtracting this, $Q^{\mathrm{SCTP}_f}$ filters out RTP-induced self-traffic and isolates only the incremental queuing after the RTP frame.

\vspace{3pt}
\noindent \textbf{Combined frame-level queuing.}
The total frame-level queuing over frame $f$'s interval is the sum of the two components:
\begin{equation}
\label{eq:5-Q-total}
    Q_f = Q_f^{\mathrm{RTP}} + Q_f^{\mathrm{SCTP}}.
\end{equation}
Multi-flow rate controller compares $Q_f$ against app-specified deadline (e.g., reduce rate when $Q_f$ exceeds it), as shown in Figure~\ref{fig:5-human-agent-workflow}.
At the SFU, which aggregates RTP traffic from multiple senders, we compute Equation~(\ref{eq:5-Q-rtp}) using the RTP frame with the highest bitrate (i.e., the frame with the longest transmission time and thus the largest queuing delay).

\begin{figure}[t]
\centering
    \includegraphics[width=1\columnwidth]{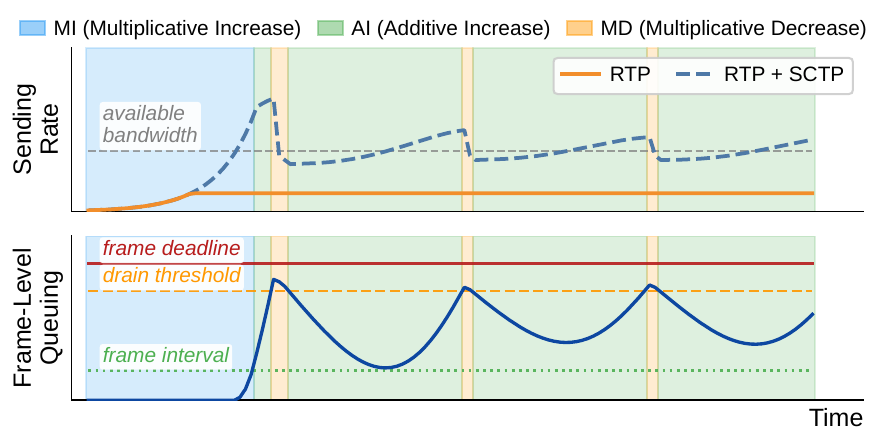}
    \vspace{-20pt}
    \caption{Frame-level queuing-based multi-flow rate control.}
\label{fig:5-multi-flow-rate-control}
\vspace{-15pt}
\end{figure}

\subsection{Frame Queuing-guided Multi-Flow Rate Controller}
\label{subsec:5-app-guided}

\JY{Given $Q_f$, we control the total sending rate for the next frame interval ($p_{f+1}$) using an MI/AI-MD mechanism to rapidly push the queuing delay closely toward (but without violating) the app-specified deadline ($D_{\mathrm{frame}}^{\mathrm{e2e}}$, typically 150~ms in video conferencing~\cite{meng2022achieving, dhawaskar2023converge}), thus ensuring maximum link utilization without degrading video quality.
Within the total sending rate, it strictly provisions the required rate for live video, and allocates the slack capacity to SCTP.
Figure~\ref{fig:5-multi-flow-rate-control} illustrates the overall rate control mechanism, composed of three regimes:
}

\begin{packeditemize} 
\item \emph{Multiplicative Increase (MI).} While the queue has not yet built up across frames ($Q_f < L$), the link has headroom and the rate grows multiplicatively: \begin{equation} 
\label{eq:5-mi-rate} 
p_{f+1} = p_f \cdot (1 + \gamma_{\mathrm{mi}}(Q_f)), \end{equation} 
where $\gamma_{\mathrm{mi}}(Q_f)$ is a function that decays monotonically as $Q_f$ grows and floors at a non-zero value near $L$ for a smooth handoff to AI (details in Appendix~\ref{appendix:5-rate-control}).  

\item \emph{Additive Increase (AI).} Once the queue persists beyond a frame interval ($L \leq Q_f < \alpha \cdot D_{\mathrm{dl}}$, \JY{where $D_{\mathrm{dl}} = D_{\mathrm{frame}}^{\mathrm{e2e}} - \mathrm{RTT}_{\min}/2$ represents the maximum allowable queuing delay}\footnote{We track the minimum RTT over the past 10 seconds, following~\cite{arun2018copa}.}), the link is near full utilization and the controller takes a fixed TCP-style AI step per frame: 
\begin{equation}
\label{eq:5-ai-rate} 
p_{f+1} = p_f + \gamma_{\mathrm{ai}} \cdot \frac{\mathrm{MSS}}{\mathrm{RTT}_{\min}} \cdot \frac{L}{\mathrm{RTT}_{\min}}, 
\end{equation} where $\mathrm{MSS}/\mathrm{RTT}_{\min}$ is the canonical per-RTT AI increment, the extra $L/\mathrm{RTT}_{\min}$ rescales it from per-RTT to our per-frame interval, and $\gamma_{\mathrm{ai}}$ is a tunable gain (Appendix~\ref{appendix:5-rate-control}).

\item \emph{Multiplicative Decrease~(MD).} 
    Once the queue approaches the deadline ($Q_f \geq \alpha \cdot D_{\mathrm{dl}}$), we anchor the rate on the \JY{receiver-observed SCTP throughput~$r_f$ (calculated from the  SCTP ACKs)} so the queue drains faster than it fills: 
    \begin{equation} 
    \label{eq:5-draining-rate} p_{f+1} = \delta \cdot r_f, \qquad \delta \leq 1. 
    \end{equation} 
    Anchoring on $r_f$ instead of a fixed multiplicative decrease of $p_f$ prevents over-correction: once $Q_f$ drops below $L$, the controller resumes MI near the available bandwidth.
\end{packeditemize}

\vspace{3pt}
\noindent
\textbf{Video rate control via agent-probed bandwidth.}
Given the observed receiver-side rate $r_f$ as an estimate of currently available bandwidth, we strictly provision the required video bitrate $b_{req}$ (e.g., 10~Mbps~\cite{dhawaskar2023converge}) to RTP, and allocate the remaining slack $p_{f+1} - b_{req}$ to SCTP (if $r_f$ falls below $b_{req}$, the entire budget is exclusively assigned to RTP). 
\REV{
We enforce the allocated RTP rate by overriding the minimum-rate parameter of the underlying CCA (e.g., GCC~\cite{carlucci2016analysis}), without modifying its internal logic so that the CCA retains its original behavior when agent flows are absent. 
}
Importantly, this naturally resolves the slow recovery problem of conventional delay-based RTP CCAs. As agent flows probe the link at rates well above $b_{req}$, bootstrapping the video rate from the observed $r_f$ drastically accelerates the video bitrate ramp-up.

\vspace{3pt} 
\noindent 
\textbf{\JY{Agent rate control with pacing.} }
\JY{
To enforce the rate allocated to SCTP, we add a pacer at the SCTP output queue.  
In our multi-flow setting, pacing the bulky context flow smooths out transient bursts, reducing their impact on the co-located video flow. Because the pacer drives the actual sending rate, we set $\mathrm{cwnd} = 2 \cdot (p_{f+1} - b_{req}) \cdot \mathrm{RTT}_{\min}$ with headroom to allow sufficient pacing rate, following BBR~\cite{cardwell2016bbr} that treats pacing as the primary rate controller.
}

\section{Multi-Agent Flow Coordinator}
\label{sec:6-coordinator-multi-agent}

\begin{figure}[t]
    \centering
        \centering
        \includegraphics[width=1\columnwidth]{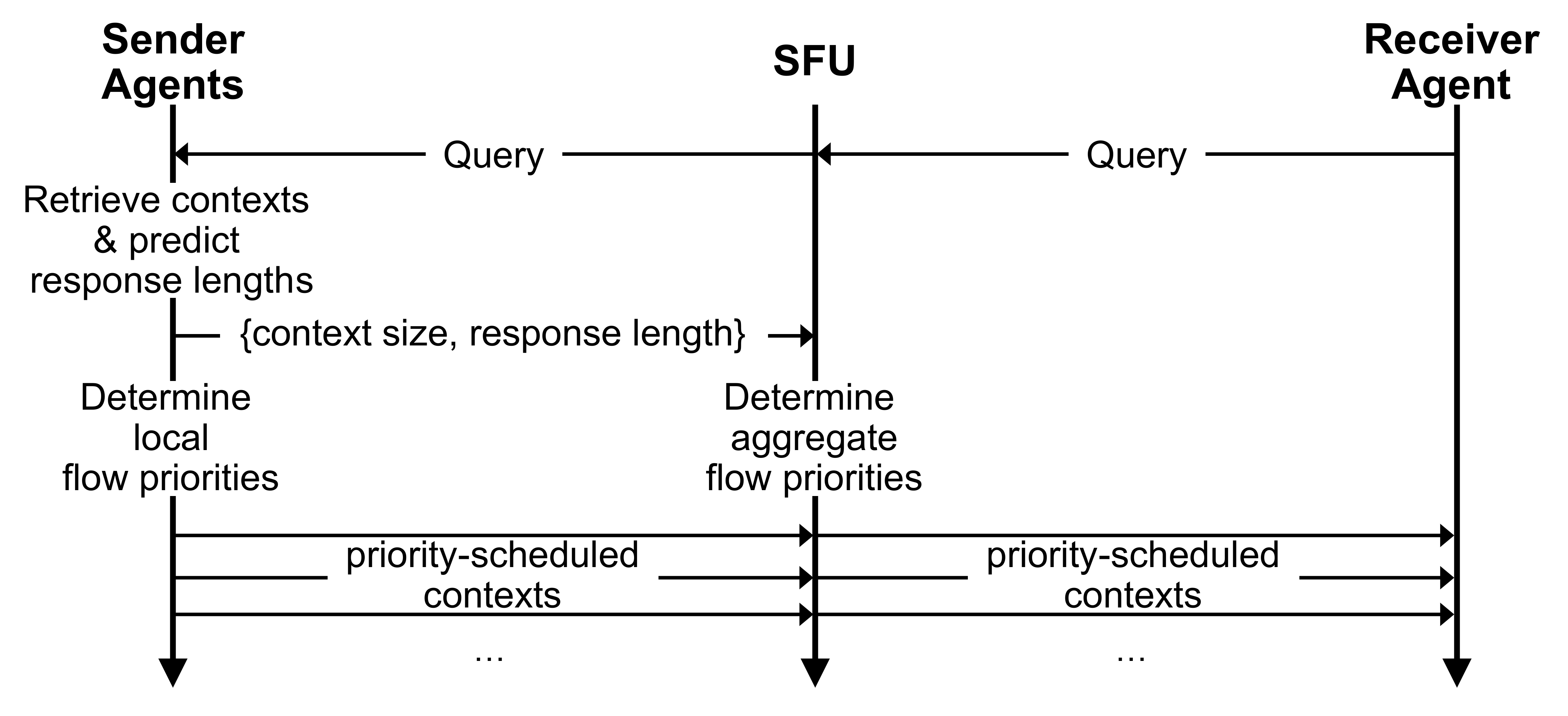}
            \vspace{-20pt}
            \caption{Multi-agent flow coordinator workflow.}
            \label{fig:6-multi-agent-workflow}
            \vspace{-15pt}
\end{figure}
%

Within the SCTP-allocated rate budget set by human-agent flow coordinator, multi-agent flow coordinator schedules transmission priorities across agent flows to minimize agent response time. 
Figure~\ref{fig:6-multi-agent-workflow} shows the operational flow. 
When a user (receiver) issues a query, it is forwarded to the sender agents via SCTP control channels. 
Each sender retrieves relevant prebuilt KV caches and predicts the response length by running a single-step LLM decoding.
Based on the KV cache sizes and available sender-SFU network bandwidths, we determine the transmission order by assigning SCTP stream priorities in a way that maximizes network-compute pipelining at the receiver's GPU (\S\ref{subsec:6-scheduling}).
We also incorporate techniques to improve robustness against network bandwidth fluctuations and response length prediction errors through dynamic priority update and opportunistic receiver-side prefill (\S\ref{subsec:6-overlapped}).

\subsection{Pipelining-aware Flow Scheduling}
\label{subsec:6-scheduling}

\noindent \textbf{Problem formulation.}
To respond to a user query, the receiver collects $N$ contexts from sender agents, performs LLM inference on each,\footnote{\REV{We assume that the compute stage processes a single decoding task at a time in FIFO order. This assumption is reasonable for single-GPU edge devices, whose limited number of cores and memory bandwidth constrain the use of batching/parallelization or preemption techniques commonly employed on datacenter GPUs~\cite{yi2020heimdall}. We discuss in \S\ref{sec:9-discussion} how our approach can be extended when more sophisticated compute-stage scheduling is available.}} and aggregates the results.
Each flow~$i$ requires network transmission time~($T_i^{net}$) followed by LLM inference time~($T_i^{comp}$), with a strict precedence: inference cannot begin until the context arrives.

Considering such constraint, the coordinator selects a transmission ordering~$\pi = (\pi_1, \pi_2, \ldots, \pi_N)$, where $1\leq \pi_{i}\leq N$ denotes the order of the $i$-th flow, to minimize the makespan:
\begin{equation}
\label{eq:6-goal}
    \pi^* = \argmin_{\pi} \; \max_{i} \; T_i^{finish}(\pi),
\end{equation}
where $T_i^{finish}(\pi)$ denotes the completion time of flow~$i$ under~$\pi$.
This is a two-machine flow shop problem~\cite{garey1976complexity}, for which Johnson's rule~\cite{johnson1954optimal} gives the optimal order.
It partitions flows into two groups: Group~A contains flows where $T_i^{net} \leq T_i^{comp}$, sorted by~$T_i^{net}$ ascending; Group~B contains the rest ($T_i^{net} > T_i^{comp}$), sorted by~$T_i^{comp}$ descending.
All Group~A flows are scheduled first, followed by Group~B.
Intuitively, Group~A flows arrive quickly and keep the compute stage busy while subsequent transfers proceed; Group~B flows, whose compute finishes fast, are placed last to shorten the tail.

\vspace{3pt}
\noindent \textbf{Workload prediction.}
Applying Johnson's rule requires predicting $T_i^{net}$ and $T_i^{comp}$ before transmission begins.
For network time, given flow size~$S_i$ and bandwidths $BW^{up}$ (sender$\to$SFU) and $BW^{down}$~(SFU$\to$receiver):
\begin{equation}
    T_i^{net} = \frac{S_i}{\min(BW^{up},\; BW^{down})},
\end{equation}
where bandwidths are estimated as moving averages of recent~(5~seconds in our settings) throughput.
For compute time, given context length~$L_i^{context}$, query length~$L_i^{query}$, and predicted output length~$\hat{L}_i^{out}$:
\begin{equation}
    T_i^{comp} = T_{prefill}(L_i^{context} + L_i^{query}) + T_{decode}(\hat{L}_i^{out}),
\end{equation}
where $T_{prefill}(\cdot)$ and $T_{decode}(\cdot)$ are offline-profiled on the target device.
\JY{
As the output length $\hat{L}_i^{out}$ is unknown before decoding, we perform a single-step LLM decoding using the prebuilt KV cache and the query to predict the response length, which has been proven feasible in recent works~\cite{jin2023s, zheng2023response}. 
Note that our scheduling is robust to prediction errors, as it only relies on the relative ordering rather than the absolute length of each response. 
For example, our evaluation in \S\ref{subsec:8-eval-microbenchmark} shows that it maintains scheduling gains even with up to a 40\% mean absolute error (MAE) in prediction. 
In practice, our predictor's error rate falls comfortably within the range, ranging from 36\% for the base LLM down to 18\% via LoRA~\cite{hu2022lora} fine-tuning with <1\% parameter overhead \gs{(detailed training methodology in Appendix~\ref{appendix:lora-training})}.
}

\subsection{Handling Prediction Errors}
\label{subsec:6-overlapped}


\noindent 
\textbf{Dynamic priority update.}
Network bandwidth fluctuations can change $T_i^{net}$. We periodically recompute flow priorities (using remaining bytes rather than the original flow sizes) when any sender-SFU bandwidth changes by more than 20\%.

\vspace{3pt}
\noindent
\textbf{Opportunistic receiver-side prefill.} 
Network bandwidth fluctuations or overestimated response lengths can leave receiver-side compute resources idle.
\REV{
To mitigate this, we send the corresponding raw contexts first (using SCTP over DTLS~\cite{rfc8261}, keeping it confidential from the RTC platform provider) whenever the receiver must fetch KV caches from the sender to answer a user's query (raw contexts are typically only a few KB, adding negligible overhead).
This allows the receiver to opportunistically prefill the contexts while its GPU is idle.
Meanwhile, the KV caches are streamed backward from the last token, allowing locally computed and transferred KV caches to meet in the middle without dependency conflicts.
}
\section{Implementation}
\label{sec:7-implementation}


\noindent \textbf{RTC framework.}
We implement \sysname{} in $\sim$11k lines of C++ atop Google WebRTC~\cite{website:WebRTC}, preserving its public API and core RTP stack.
Four hooks (two in GCC's bandwidth estimators and two in dcSCTP) feed one-way delay, acked bytes, SCTP chunk send timestamps, and ACK RTT samples to the human--agent flow coordinator, which sets the pacing rate (enforced by a token-bucket pacer~\cite{wikipedia_token_bucket}) and minimum RTP bitrate.
The multi-agent flow coordinator reuses WebRTC data channels for context streaming and tracks their state; a \sysname{}-managed SCTP control channel carries queries, context metadata, and flow-priority decisions.
Our SFU uses the same build to relay RTP/SCTP, prioritize the main speaker's video~\cite{lin2022gso}, and route queries to designated senders via matching SCTP streams.


\vspace{3pt}
\noindent \textbf{AI agent.}
We use llama.cpp~\cite{llamacpp} through its public C API.
Because the API only supports whole-state KV-cache dump/restore, our thin C++ wrapper (\texttt{LLMEngine}, $\sim$3k lines) adds in-memory load/save and KVZip-compatible per-layer sparse variants~\cite{kim2025kvzip}, feeding SCTP payloads directly to the model without disk I/O.
\section{Performance Evaluation}
\label{sec:8-eval}

Our key evaluation results are summarized as follows.

\begin{packeditemize}
    \item \textbf{Performance improvements.} \sysname{} improves video quality by up to 1.5$\times$ and reduces agent response latency by 31\% compared to the baselines (\S\ref{subsec:8-eval-real-world}, \S\ref{subsec:8-eval-varying}).
    \item \textbf{Generality.} \sysname{} achieves consistent performance gains across diverse network conditions, LLM models, video bitrates, and number of concurrent agent flows (\S\ref{subsec:8-eval-varying}).
    \item \textbf{Microbenchmarks.} We present the effectiveness of each component of \sysname{} through ablation studies~(\S\ref{subsec:8-eval-microbenchmark}).
    \item \textbf{Low resource overheads.} \sysname{} incurs negligible overheads in power consumption, CPU, and memory (\S\ref{subsec:8-eval-deep-dive}).
\end{packeditemize}

\subsection{Evaluation Setup}
\label{subsec:8-setup}

\noindent
\textbf{Datasets.}
We generate 109 multi-document QA sets from MuSiQue~\cite{trivedi2022musique} and 75 meeting summarization sets from QMSum~\cite{zhong2021qmsum}.
Each set contains three agent contexts, with $<$4k tokens per context for QA and 8--16k for summarization.
Full per-model decoding-length statistics are in Appendix~\ref{appendix:dataset-stats}.
We compress the resulting KV caches to 30\% using KVZip~\cite{kim2025kvzip}.
Unless specified otherwise, we report performance on summarization.

\vspace{3pt}
\noindent
\textbf{LLMs.} 
We use three representative models: Qwen2.5-3B, Qwen2.5-7B~\cite{bai2025qwen2}, and Llama3.1-8B~\cite{dubey2024llama}, all using W4A16 quantization. Qwen2.5-3B is used unless specified otherwise.

\begin{figure}[t]
  \centering
  \begin{minipage}[t]{.37\textwidth}
    \centering
    \includegraphics[width=\textwidth]{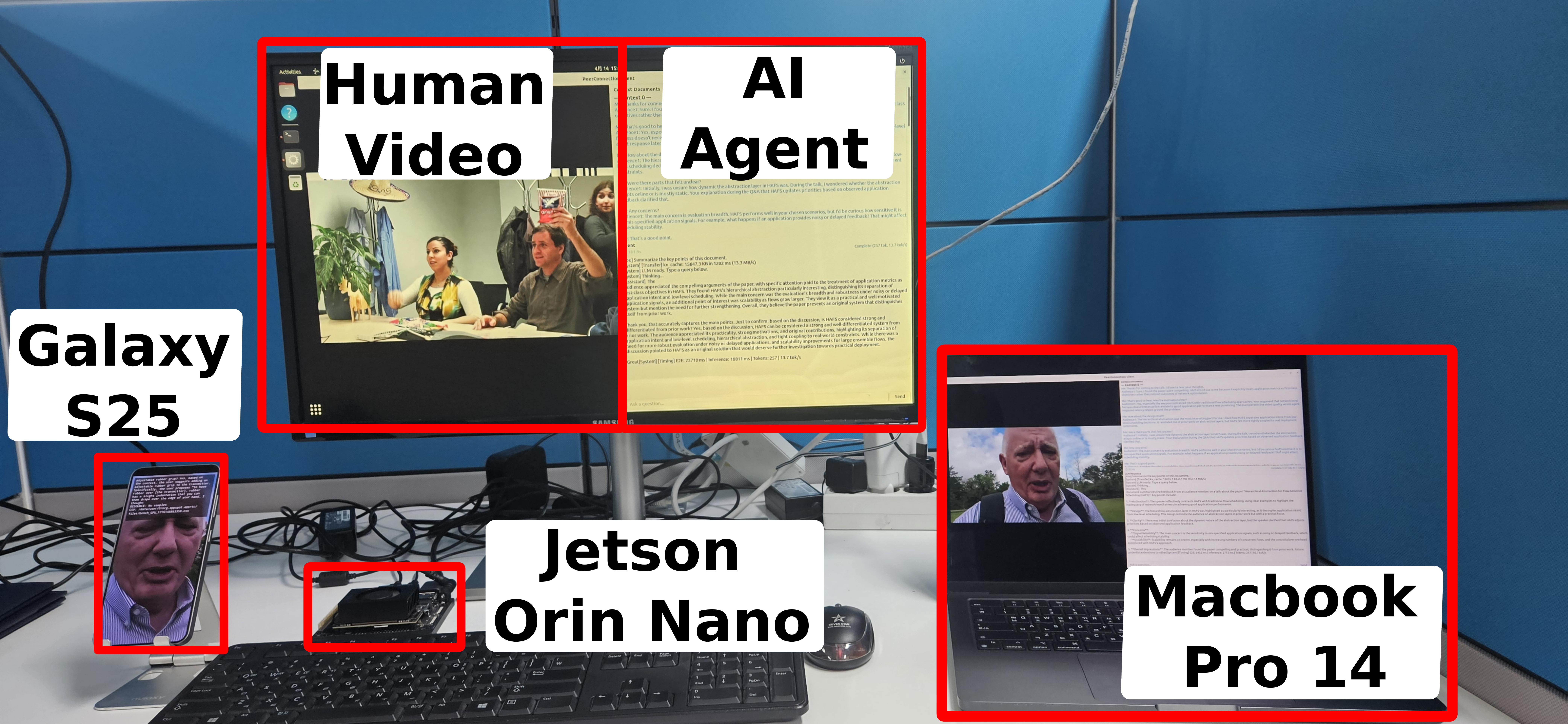}
    \vspace{-20pt}
    \caption{\sysname{} testbed.}
    \label{fig:7-testbed}
  \end{minipage}%
  \vspace{-15pt}
\end{figure}

\vspace{3pt}
\noindent
\textbf{Experiment methodology.} 
\begin{packeditemize}
    \item \textbf{Real-world testbed (Figure~\ref{fig:7-testbed}).}
    Our in-lab testbed comprises three edge devices (NVIDIA Jetson Orin Nano 8~GB~\cite{NVIDIA2025JetsonOrinnx}, MacBook Pro 14~(M3 Pro, 18~GB RAM)~\cite{wikipedia_macbook_pro_apple_silicon}, and Samsung Galaxy S25~(Qualcomm Adreno 830 GPU and Hexagon NPU)) and a Linux desktop SFU (i7-12700 CPU, 32~GB RAM).
    The edge devices use the Xiaomi BE 3600 Wi-Fi~6 AP's 5~GHz link, with $\approx$30~ms wired delay between the AP and SFU.
    Each run pairs two devices as sender (uploading human video and agent KV caches) and receiver.
    We repeat each configuration for one hour during daytime, with contention from co-channel Wi-Fi networks in the building.

    \item \textbf{Trace-driven emulation.}
    For repeatable experiments, we launch peer processes on an Azure VM (Ubuntu 22.04.5, AMD EPYC 9V84 CPU with 98 logical CPUs, 866~GiB RAM) and reshape link bandwidth with \texttt{tc} using Wi-Fi~6 (257.79$\pm$19.25 Mbps) and 5G (482.95$\pm$228.49 Mbps) traces from our testbed and prior work~\cite{meng2022achieving}. We emulate inference by inserting delays matching the inference latencies profiled on our testbed devices.
\end{packeditemize}

\vspace{3pt}
\noindent \textbf{Baselines.} We compare \sysname{} with the following schemes:
\begin{packeditemize}
    \item \textsc{Raw} runs both the prefill and decode phases entirely on the receiver device with raw contexts.
    \item \textsc{No Coordination (NC)} is a default scheme, identical to WebRTC and Zoom, which runs RTP and SCTP independently without any coordination, and also applies round-robin scheduling for agent flows.
    \item \textsc{Conservative Coordination (CC)} emulates the SCTP throttling behavior in Teams, where we send 20~MB in 5~seconds at a constant rate. It also applies round-robin scheduling across agent flows.
    \item \textsc{FSE (Flow State Exchange)~\cite{islam2022real}} is a transport-level cross-flow optimization technique that aggregates bandwidth estimates from RTP and SCTP and then allocates rates fairly. This does not recognize app-level objectives, such as video delay or priority between multi-agent flows.
\end{packeditemize}

\begin{figure}[t]
  \centering
  \begin{minipage}[t]{0.48\textwidth}
      \begin{subfigure}[t]{0.48\textwidth}
        \centering
        \includegraphics[width=\textwidth]{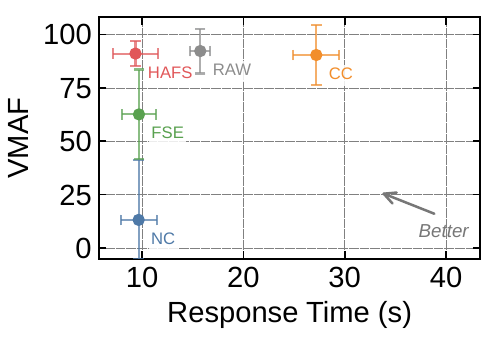}
        \vspace{-15pt}
        \subcaption{QA.}\label{fig:real-mbp-qa}
      \end{subfigure}%
      \hfill
      \begin{subfigure}[t]{0.48\textwidth}
        \centering
        \includegraphics[width=\textwidth]{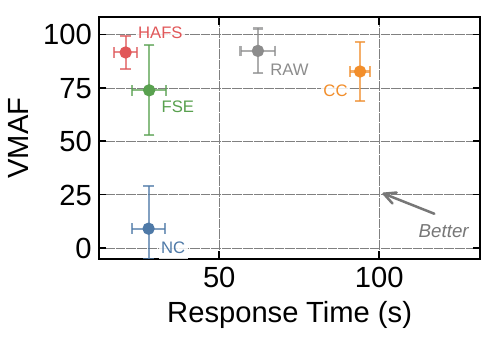}
        \vspace{-15pt}
        \subcaption{Summarization.}\label{fig:real-mbp-summ}
      \end{subfigure}
      \vspace{-10pt}
      \caption{
      Real-world app performance on MacBook Pro.}\label{fig:real-mbp}
      \vspace{-10pt}
  \end{minipage}
\end{figure}

\begin{figure}[t]
  \centering
  \begin{minipage}[t]{0.48\textwidth}
      \begin{subfigure}[t]{0.48\textwidth}
        \centering
        \includegraphics[width=\textwidth]{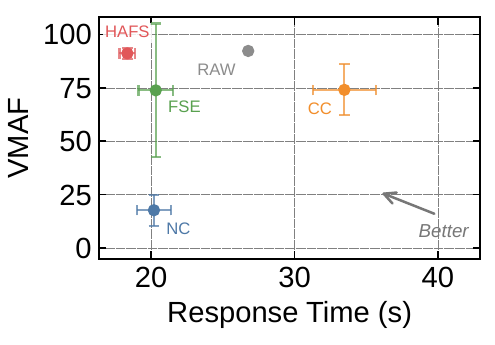}
        \vspace{-15pt}
        \subcaption{QA.}\label{fig:real-jet-qa}
      \end{subfigure}%
      \hfill
      \begin{subfigure}[t]{0.48\textwidth}
        \centering
        \includegraphics[width=\textwidth]{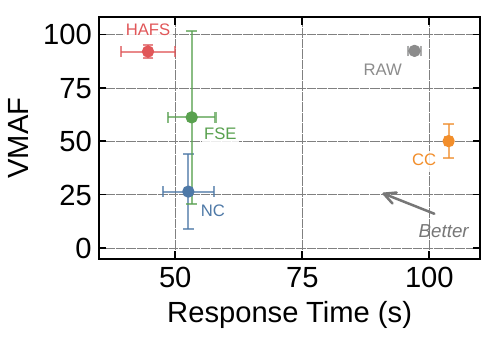}
        \vspace{-15pt}
        \subcaption{Summarization.}\label{fig:real-jet-summ}
      \end{subfigure}
      \vspace{-10pt}
      \caption{Real-world app performance on Jetson Orin.}\label{fig:real-jet}
      \vspace{-10pt}
    \end{minipage}
\end{figure}

\subsection{Real-World Evaluation}
\label{subsec:8-eval-real-world}

We first run all schemes end-to-end on the testbed, covering both scenarios on each edge device. Figure~\ref{fig:real-mbp} plots VMAF against agent response time.
\sysname{} lands at (91, 9\,s) for QA and (92, 21\,s) for Summarization. This matches with \textsc{Raw}'s visual quality while shrinking its response time by 1.7$\times$ and 3.0$\times$, respectively. Other baselines sacrifice one axis: \textsc{NC} finishes fast but collapses VMAF to 13 and 9, \textsc{CC} preserves quality at the cost of 2.9--4.5$\times$ response time penalty, and \textsc{FSE} closes the latency gap versus \textsc{NC} but still trails \sysname{} by 17--29 VMAF because its rate control is blind to video deadlines. Figure~\ref{fig:real-jet} shows the same trend on Jetson Orin: its slower GPU raises every scheme's absolute response time, but the ordering across schemes is preserved.

\subsection{Performance over Varying Environments}
\label{subsec:8-eval-varying}

\begin{figure}[t]
  \centering
  \begin{minipage}[t]{0.48\textwidth}
    \begin{minipage}[t]{0.48\textwidth}
      \centering
      \includegraphics[width=\textwidth]{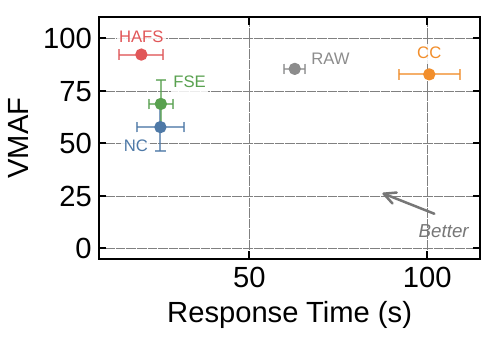}
      \vspace{-15pt}
      \subcaption{Wi-Fi~6.}\label{fig:app-perf-net-w6}
    \end{minipage}%
    \hfill
    \begin{minipage}[t]{0.48\textwidth}
      \centering
      \includegraphics[width=\textwidth]{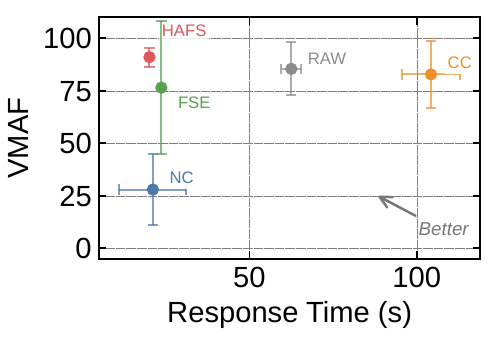}
      \vspace{-15pt}
      \subcaption{5G.}\label{fig:app-perf-net-5g}
    \end{minipage}
    \vspace{-10pt}
    \caption{Performance under varying network conditions.}\label{fig:8-app-perf-net}
  \end{minipage}
  \vspace{-10pt}
\end{figure}

\vspace{3pt}
\noindent
\textbf{Network conditions.} Figure~\ref{fig:8-app-perf-net} plots VMAF against response time under emulated Wi-Fi and 5G traces. 
\sysname{} achieves 91 VMAF at 19~s under Wi-Fi and 87 VMAF at 20\,s under the more variable 5G trace, while \textsc{NC} and \textsc{FSE} sacrifice VMAF under bursty 5G and \textsc{CC} needs 100+~s to recover.

\begin{table}[t]
    \centering
    \caption{Network statistics under a Wi-Fi~6 trace.}
    \vspace{-5pt}
    \label{tab:bufferbloat}
    \small
    \setlength{\tabcolsep}{3pt}
    \begin{tabular}{lrrrrr}
    \toprule
     & \multicolumn{3}{c}{Video} & \multicolumn{2}{c}{SCTP} \\
    \cmidrule(lr){2-4}\cmidrule(lr){5-6}
    Alg.
      & \shortstack{TP\\(Mbps)}
      & \shortstack{Delay\\p50/p95 (ms)}
      & \shortstack{Stall\\{>}150\,ms}
      & \shortstack{TP\\(Mbps)}
      & \shortstack{RTT\\p50/p95 (ms)} \\
    \midrule
    \textsc{Raw}  & 8.19 & 46 / 89   &  0.0\% &   0.9 &  30 / 30  \\
    \textsc{NC}   & 3.66 & 46 / 318  & 25.7\% & 207.6 & 157 / 432 \\
    \textsc{FSE}  & 7.05 & 60 / 291  & 24.4\% & 196.8 & 156 / 308 \\
    \textsc{CC}   & 8.15 & 43 / 85   &  0.1\% &  56.1 &  30 / 58  \\
    \sysname{}    & 8.16 & 47 / 107  &  0.1\% & 192.5 &  82 / 94  \\
    \bottomrule
    \end{tabular}
    \vspace{-5pt}
\end{table}

Table~\ref{tab:bufferbloat} drills into Wi-Fi~6 statistics. Over
\textsc{NC}, \sysname{} delivers 2.2$\times$ higher video throughput,
3.0$\times$ lower p95 frame delay, and near-zero stalls
(0.1\% vs.\ 25.7\%). The differentiator is not context bandwidth
(comparable across schemes) but 4.6$\times$ lower SCTP p95 RTT
(94 vs.\ 432\,ms): our multi-flow rate control keeps queues
bounded under video deadlines.

\begin{figure}[t]
  \centering
  \begin{minipage}[t]{0.48\textwidth}
    \begin{minipage}[t]{0.48\textwidth}
      \centering
      \includegraphics[width=\textwidth]{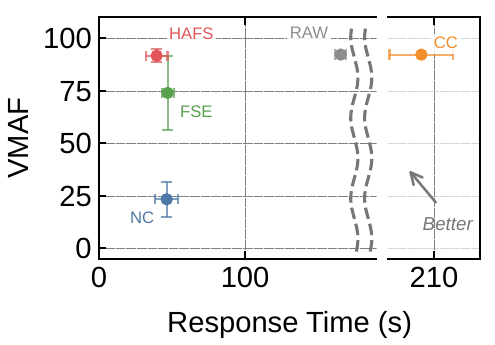}
      \vspace{-15pt}
      \subcaption{Qwen-2.5 7B.}\label{fig:app-perf-llm-7b}
    \end{minipage}%
    \hfill
    \begin{minipage}[t]{0.48\textwidth}
      \centering
      \includegraphics[width=\textwidth]{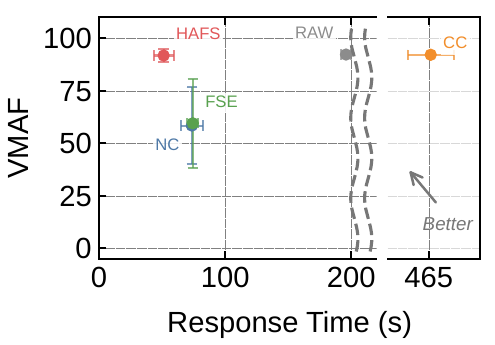}
      \vspace{-15pt}
      \subcaption{Llama3.1-8B.}\label{fig:app-perf-llm-llama}
    \end{minipage}
    \vspace{-10pt}
    \caption{Performance over varying LLM models.}\label{fig:8-app-perf-llm}
  \end{minipage}
  \vspace{-15pt}
\end{figure}

\vspace{3pt} 
\noindent 
\textbf{LLM models.} Across Qwen2.5-7B and Llama3.1-8B, \sysname{} reduces response time by 15--16\% and 30--31\% relative to \textsc{NC}/\textsc{FSE}, and is 4.2$\times$/3.8$\times$ faster than \textsc{Raw} (Figure~\ref{fig:8-app-perf-llm}).

\begin{figure}[t]
  \centering
  \begin{minipage}[t]{0.48\textwidth}
      \begin{minipage}[t]{.48\textwidth}
        \centering
        \includegraphics[width=\textwidth]{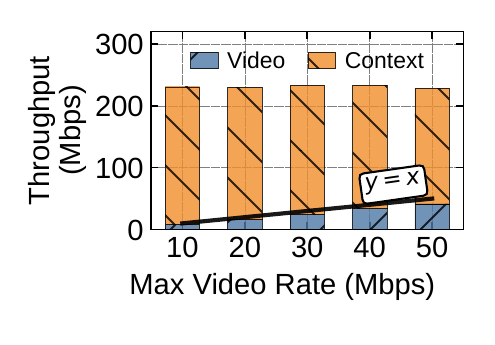}
        \vspace{-25pt}
        \caption{\sysname{} with higher RTP requirements.}\label{fig:8-eval-microbenchmark-rtp}
      \end{minipage}
      \hfill
      \begin{minipage}[t]{.48\textwidth}
        \centering
        \includegraphics[width=\textwidth]{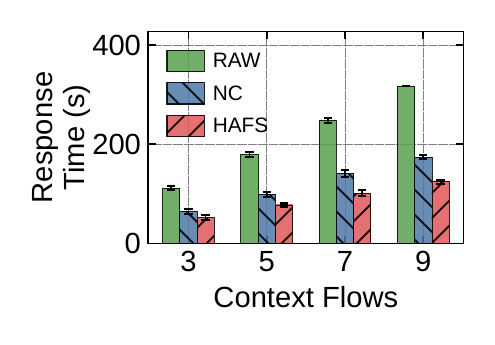}
        \vspace{-25pt}
        \caption{\sysname{} with varying concurrent agent flows.}\label{fig:8-eval-microbenchmark-flows}
      \end{minipage}%
  \end{minipage}
  \vspace{-15pt}
\end{figure}

\vspace{3pt} 
\noindent 
\textbf{Video bitrates.} Figure~\ref{fig:8-eval-microbenchmark-rtp} sweeps the maximum video bitrates from 10 to 50\,Mbps considering emerging high-bitrate RTC scenarios~(e.g., volumetric video~\cite{guan2023metastream, lee2025deltastream}). We transfer a bulk context flow to saturate the rest of the link. \sysname{} tracks the video cap within 2~Mbps (10$\to$8.3, 50$\to$41.3~Mbps; the remaining gap is encoder-side), and the context flow absorbs the residual capacity (222$\to$186~Mbps).

\vspace{3pt}
\noindent
\textbf{Number of concurrent agent flows.}
With Llama3.1-8B, increasing concurrent context flows from 3 to 9 widens \sysname{}'s response-time advantage over \textsc{NC} from 53 to 64~s and over \textsc{Raw} from 84 to ${\approx}480$~s (Figure~\ref{fig:8-eval-microbenchmark-flows}).
More contexts let Johnson's rule better overlap network transfer with GPU inference.

\subsection{Microbenchmarks}
\label{subsec:8-eval-microbenchmark}

\begin{figure}[t]
  \centering
  \begin{minipage}[t]{0.48\textwidth}
      \begin{minipage}[t]{.48\textwidth}
        \centering
        \includegraphics[width=\textwidth]{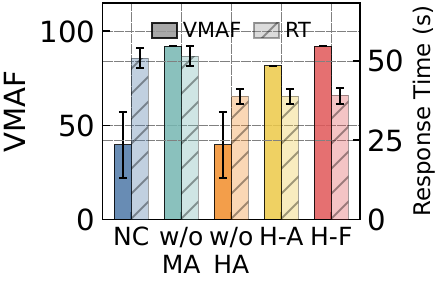}
        \vspace{-20pt}
        \caption{Ablation study on \sysname{} module.}\label{fig:8-eval-microbenchmark-ablation}
      \end{minipage}%
      \hfill
      \begin{minipage}[t]{.48\textwidth}
        \centering
        \includegraphics[width=\textwidth]{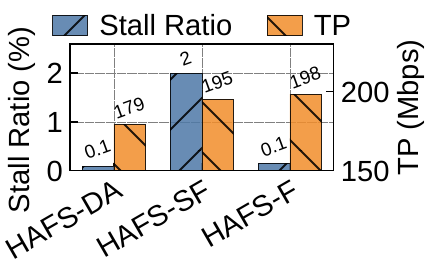}
        \vspace{-20pt}
        \caption{Ablation study on human-agent coordinator.}\label{fig:8-eval-microbenchmark-ablation-multi}
      \end{minipage}%
  \end{minipage}
  \vspace{-15pt}
\end{figure}

\noindent\textbf{Ablation study.} Figure~\ref{fig:8-eval-microbenchmark-ablation} shows that removing multi-agent coordinator (w/o~MA) preserves VMAF but increases response time, while removing human-agent coordinator (w/o~HA) lowers VMAF but preserves response time. Enabling both without video rate control (H-A) recovers response time gain but trails w/o~MA by a few VMAF points due to GCC's conservative re-probing. Full \sysname{} (H-F) closes this gap, improving VMAF by 19\%.

\vspace{3pt}
\noindent
\textbf{Human-agent flow coordinator.} Figure~\ref{fig:8-eval-microbenchmark-ablation-multi} shows that deadline-agnostic \textsc{HAFS-DA}, which reduces rates when per-frame queuing exceeds the frame interval, achieves a 0.1\% frame stall ratio at a 10\% throughput cost. Using only a single RTCP feedback (\textsc{HAFS-SF}) delays congestion response, yielding a 2\% stall ratio. \textsc{HAFS-F}, the full coordinator, combines a low stall ratio with high throughput.

\begin{figure}[t]
  \centering
  \begin{minipage}[t]{.23\textwidth}
    \centering
    \includegraphics[width=\textwidth]{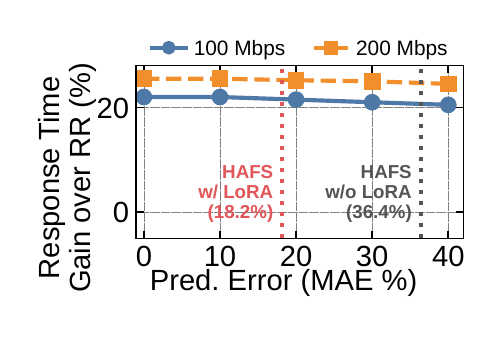}
    \vspace{-20pt}
    \caption{Impact of LLM latency prediction error.
    }
    \label{fig:8-eval-microbenchmark-error}
  \end{minipage}%
  \hfill
  \begin{minipage}[t]{.23\textwidth}
    \centering
\includegraphics[width=\textwidth]{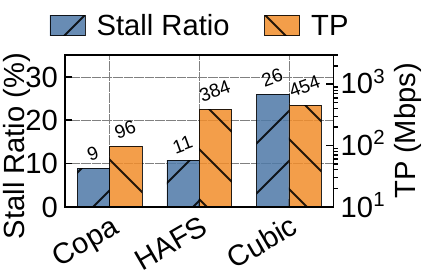}
    \vspace{-20pt}
    \caption{Comparison with various CCAs.}\label{fig:8-eval-deep-dive-cca}
  \end{minipage}%
  \vspace{-15pt}
\end{figure}

\vspace{3pt}
\noindent
\textbf{\REV{Multi-agent flow coordinator.}} 
Refer to Appendix~\ref{appendix:multi-agent-microbenchmark} for operation timeline example and scheduling policy comparison.

\vspace{3pt}
\noindent
\textbf{Impact of LLM latency prediction error.} Figure~\ref{fig:8-eval-microbenchmark-error} shows that \sysname{} outperforms round-robin across bandwidths with errors manually injected into ground truth, because its multi-agent coordinator needs only relative flow ordering, not absolute lengths. 
Although LoRA fine-tuning reduces Qwen-2.5-3B's mean absolute error~(MAE) from 36.4\% to 18.2\%, the base LLM preserves ordering well enough to match LoRA-level scheduling performance.

\subsection{\sysname{} Deep Dive}
\label{subsec:8-eval-deep-dive}

\noindent
\textbf{Comparison with various CCAs.}
Figure~\ref{fig:8-eval-deep-dive-cca} evaluates concurrent video streaming and bulky KV-cache transfers over fluctuating 5G traces, with the transfers controlled by \sysname{}, CUBIC, or Copa.
Loss-based CUBIC achieves high throughput but allows queue buildup that delays video frames (26\% frame stall ratio).
Copa's delay-based control yields fewer stalls (9\%) but only 21\% of CUBIC's throughput in this setting.
By accounting for video frame deadlines, \sysname{} achieves near-CUBIC throughput and near-Copa stall ratio.
\REV{
Refer to Appendix~\ref{appendix:ablation-cca} for app-level comparisons.
}

\begin{figure}[t]
  \centering
  \begin{minipage}[t]{0.23\textwidth}
    \centering
    \includegraphics[width=\textwidth]{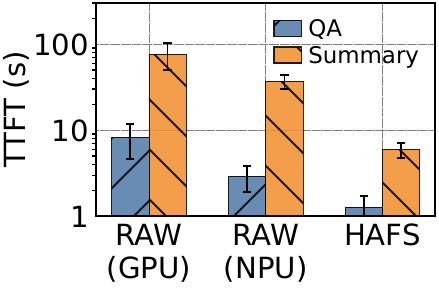}
    \vspace{-20pt}
    \captionof{figure}{Time-to-first-token on the S25 device.}
    \label{fig:7-ttft}
  \end{minipage}
  \hfill
  \begin{minipage}[t]{0.23\textwidth}
    \centering
    \includegraphics[width=\textwidth]{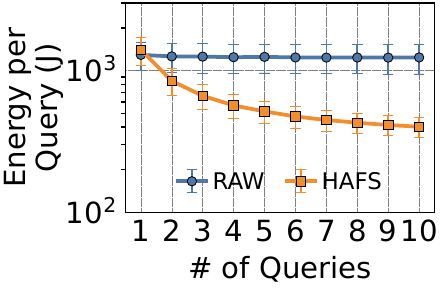}
    \vspace{-20pt}
    \captionof{figure}{Energy per query on the same context.}
    \label{fig:7-energy}
  \end{minipage}
  \vspace{-10pt}
\end{figure}

\vspace{3pt} 
\noindent 
\textbf{Comparison with NPU prefill baseline.} 
Figure~\ref{fig:7-ttft} compares the time-to-first-token~(TTFT) of QA and summarization tasks on the Galaxy S25 of \sysname{} and \textsc{Raw} with GPU backend (llama.cpp) and an NPU backend (Qualcomm Genie).
\sysname{} achieves 6.4$\times$/12.8$\times$ and 2.2$\times$/6.2$\times$ TTFT speedups over the GPU and NPU, suggesting that remote KV-cache streaming is beneficial in current network bandwidths.


\vspace{3pt} 
\noindent \textbf{Energy consumption.} Figure~\ref{fig:7-energy} reports Jetson's per-query energy for $N$ consecutive summarization queries. \textsc{Raw} repeats full local prefill, consuming 1240\,J per query. \sysname{} amortizes one prefill per session: it consumes more at $N{=}1$ (1404\,J) but less than \textsc{Raw} from $N{=}2$, reaching 402\,J at $N{=}10$ (3.1$\times$ savings).

\begin{table}[t]
  \centering
  \small
  \setlength{\tabcolsep}{3pt}  
  \vspace{-3pt}
  \caption{System-level overhead comparison.}
  \vspace{-5pt}
  \label{tab:8-eval-overhead}
  \begin{tabular}{l cc cc cc}
    \toprule
    & \multicolumn{2}{c}{\textbf{MacBook}} & \multicolumn{2}{c}{\textbf{Jetson}} & \multicolumn{2}{c}{\textbf{S25}} \\
    \cmidrule(lr){2-3} \cmidrule(lr){4-5} \cmidrule(lr){6-7}
    & \textsc{Raw} & \sysname{} & \textsc{Raw} & \sysname{} & \textsc{Raw} & \sysname{} \\
    \midrule
    \textbf{CPU (\%)}  & 27.5 & 30.1 & 42.0 & 39.8 & 60.4 & 55.8 \\
    \textbf{GPU (\%)}  & 46.9 & 33.7 & 50.4 & 32.4 & 64.1 & 27.4 \\
    \textbf{Mem (MB)}  & 2467 & 2585 & 1998 & 1638 & 3292 & 3188 \\
    \bottomrule
  \end{tabular}
  \vspace{-15pt}
\end{table}

\vspace{3pt}
\noindent
\textbf{System-level overheads.}
Table~\ref{tab:8-eval-overhead} shows that \sysname{} incurs negligible overheads over \textsc{Raw}: CPU utilization differs by less than 5 percentage points and memory usage by at most 20\% across devices. GPU utilization drops by 13--37 percentage points as network transfer replaces part of local prefill.
\section{Discussion and Future Work}
\label{sec:9-discussion}

\noindent{\bf Generality.}
\REV{
\sysname{} optimizes offline sender-side prefill and KV cache streaming pipeline for common collaborative tasks such as document review, focusing on settings where multiple peers repeatedly query a shared context and sufficient network bandwidth is available (\S\ref{subsec:3-base-pipeline}).
The optimal pipeline may vary with collaboration patterns and available network/compute resources; for example, semantically similar peer queries may favor sender-side decoding and response caching.
}

\vspace{3pt}
\noindent{\bf Context processing granularity.}
\REV{
We assume that contexts are prefilled into KV caches at a fixed, task-specific granularity (1--4K tokens for QA and 8--16K for summarization), each context representing an individual analysis unit such as conversation transcript or document section.
In practical deployment, determining the optimal unit for partitioning the input context is critical: excessively coarse-grained partition results in resource wastage from transmitting query-irrelevant tokens, whereas overly fine-grained partitioning causes accuracy drop due to missing cross-attention across adjacent context chunks. 
We plan to investigate KV cache granularity optimization and selective fusion techniques~\cite{yao2025cacheblend} in future.
}

\vspace{3pt}
\noindent{\bf Scalability}.
As the number of users in the RTC session grows, memory constraints on per-flow transport buffers may limit SCTP's congestion window and reduce throughput, especially at the SFU, where all senders' contexts converge.
Recent LLM serving systems address similar pressure by offloading KV caches to hierarchical storage~\cite{gao2024cost}.
We plan to integrate these techniques into \sysname{} to improve scalability.

\vspace{3pt}
\noindent{\bf \REV{Advanced LLM inference scheduling}}.
\REV{
\sysname{} currently assumes single-task, FIFO decoding on a resource-constrained edge GPU (\S\ref{subsec:6-scheduling}).
We plan to incorporate the following advanced LLM inference scheduling mechanisms when resource constraints are relaxed.
(i) If independent tasks can run concurrently on GPU partitions (e.g., as Streaming Multiprocessor partitioning on NVIDIA GPUs~\cite{bakita2023hardware}) or on both GPU and NPU, Equation~\ref{eq:6-goal} can be modeled as an NP-hard hybrid flow shop problem~\cite{ruiz2010hybrid}, motivating efficient scheduling heuristics that account for system-level effects (e.g., mobile SoC shared memory bandwidth contention).
(ii) With sufficient GPU memory, decoding for a newly arrived KV cache can be batched with ongoing decoding to reduce overall latency (e.g., using iteration-level batching~\cite{yu2022orca}).
In this case, each flow's decoding latency also depends on the evolving batch size and constituent KV cache sizes, requiring batch-aware latency modeling and scheduling.
(iii) If memory bandwidth permits low-overhead preemption, \sysname{} can incorporate priority scheduling across concurrent user queries according to their latency requirements.
}

\vspace{3pt}
\noindent{\bf Limitations}. 
\sysname{} jointly controls RTP/SCTP to maximize SCTP throughput within RTP deadlines.
When other hosts' traffic causes queuing, it reduces SCTP's rate to prioritize RTP rather than probing for fair share~(Appendix~\ref{appendix:fairness}).
End-host scheduling alone cannot enforce cross-host priorities; additional in-network support is required.

\section{Related Work}
\label{sec:10-related-work}

\noindent{\bf Real-time communication (RTC)}. Many techniques have been proposed to enhance the performance of RTC apps. 
End-host solutions, such as congestion control~\cite{wang2024pudica, salsify_nsdi18, lee2021demystifying} or loss recovery techniques~\cite{meng2024hairpin, an2025tooth}, struggle to handle sudden delay fluctuations from varying network conditions. 
However, these works primarily focus on a single video flow scenario, whereas \gs{on-device} agent-augmented RTC requires efficient handling of co-existing bulky agent traffic flows.

\vspace{3pt}
\noindent{\bf Agent communication frameworks}.
Recent agent-to-agent protocols (e.g., Google A2A~\cite{a2a},
IBM ACP~\cite{acp}, Cisco ANP~\cite{anp}) primarily define
interoperable interfaces across vendors (e.g., A2A's JSON-based agent card describing identities and capabilities). 
They target text-centric, single-flow apps, whereas agent-augmented RTC involves concurrent high-volume flows from humans and agents.

\vspace{3pt}
\noindent{\bf Agent serving frameworks}. 
Several recent systems also focus on \gs{fast} LLM and agent inference serving in datacenters.
Many systems reduce redundant prefill overhead by streaming and fusing KV caches across nodes~\cite{zhong2024distserve, luo2025autellix, yao2025cacheblend, liu2024cachegen}. 
However, these systems target high-bandwidth environments ($>$100~Gbps Ethernet) and assume homogeneous traffic, neither of which holds for on-device agent-augmented RTC, where agent flows must coexist with latency-sensitive video traffic under constrained wireless links.
\section{Conclusion}

We presented \sysname{}, a system for emerging agent-augmented RTC apps to guarantee both high live video streaming quality and low agent response latency.
\sysname{} coordinates concurrent human and agent flows through an app-guided multi-flow transport, which leverages \gs{heterogeneous flows' app-layer semantic requirements} to make fine-grained scheduling decisions while also enabling easy deployment in commercial RTC frameworks.
Evaluation shows that \sysname{} significantly outperforms baselines, \gs{achieving 1.5$\times$ higher video quality and 31\% lower agent response latency.}

\balance
\clearpage
\bibliographystyle{plain}
\balance
\bibliography{hafs}

\clearpage
\nobalance
\appendix

\section*{Appendices}

\begin{figure}[t]
    \centering
    \includegraphics[width=.96\columnwidth]{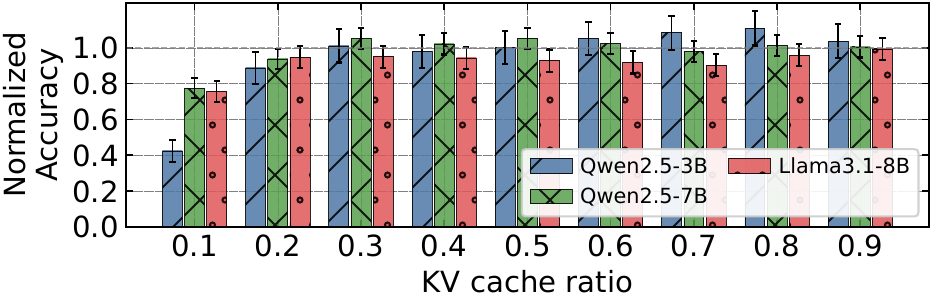}
    \caption{KVZip normalized accuracy on multi-hop QA (MuSiQue, 300 contexts, max(F1, inclusion)).}
    \label{fig:appendix-kvzip-qa}
\end{figure}

\begin{figure}[t]
    \centering
    \includegraphics[width=.96\columnwidth]{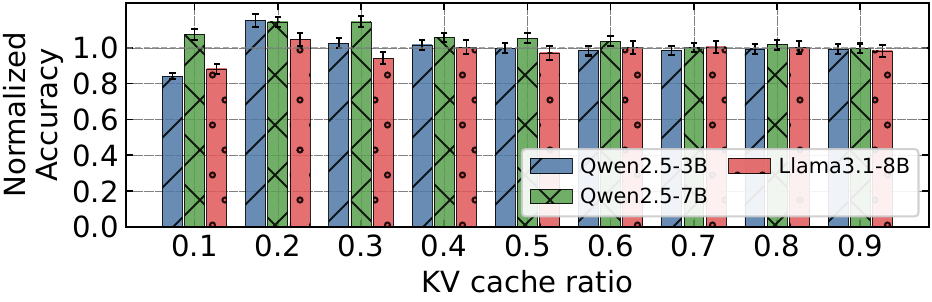}
    \caption{KVZip normalized accuracy on meeting summarization (QMSum, 300 contexts, ROUGE-L metric).}
    \label{fig:appendix-kvzip-summ}
\end{figure}

\section{KVZip Compression vs.\ Accuracy}
\label{appendix:kvzip-accuracy}

We evaluate how KVZip~\cite{kim2025kvzip} KV cache compression affects the accuracy of three on-device LLMs: Qwen2.5-3B, Qwen2.5-7B, and Llama3.1-8B that use W4A16 quantization (4-bit weights, FP16 activations and KV caches) from Hugging Face.
We sweep the KV cache retention ratio from 0.1 to 0.9 on 300 contexts each from the MuSiQue multi-hop QA dataset~\cite{trivedi2022musique} (1K--4K tokens) and the QMSum meeting summarization dataset~\cite{zhong2021qmsum} (8K--16K tokens).
Each bar in Figures~\ref{fig:appendix-kvzip-qa} and~\ref{fig:appendix-kvzip-summ} shows the accuracy at a given retention ratio normalized by the full (uncompressed) KV cache accuracy, following KVZip's evaluation protocol.

\vspace{3pt}
\noindent
\textbf{QA task} (Figure~\ref{fig:appendix-kvzip-qa}).
At a retention ratio of 0.3 (i.e., 70\% of the KV cache is evicted), Qwen2.5-3B, Qwen2.5-7B, and Llama3.1-8B all retain $\geq$95\% of their full-cache accuracy.

\vspace{3pt}
\noindent
\textbf{Summarization task} (Figure~\ref{fig:appendix-kvzip-summ}).
All three models maintain $\geq$85\% normalized ROUGE-L even at the most aggressive compression (ratio 0.1).
Summarization is inherently more robust to KV cache pruning because generating a coherent summary only requires the overall context gist, not token-level precision.

\vspace{3pt}
\noindent
\textbf{Implication for \sysname{}.}
These results validate the KV cache ratio of 0.3 used in our experiments (\S\ref{subsec:8-setup}): at this ratio, KVZip reduces network transfer by 70\% with negligible accuracy loss for the majority of models and both task types, enabling \sysname{} to deliver real-time LLM inference on resource-constrained edge devices.

\section{Measurement Studies on Commercial RTC Apps}
\label{appendix:zoom-teams}

In \S\ref{subsubsec:3-human-agent-flow-contention}, we show the measurement details of commercial RTC platforms during concurrent video and agent context transmission (i.e., large file). Through this, we report the underlying transport-layer choices revealed by our Wireshark captures.

\vspace{3pt}
\noindent
\textbf{Zoom --- bulk TCP alongside RTP.}
Zoom carries the human video flow as RTP on UDP~(the Zoom media port, 8801), and transports the agent context flow over a separate TCP connection on port 443 to its cloud server.
In our captures, the RTP flow rate is reduced from 3~to 1~Mbps by an aggressive TCP file-transfer connection. This reaches up to 200~Mbps within three seconds of the handshake completing and sustaining a standing queue until the transfer terminates.

\begin{figure}[t]
    \centering
    \includegraphics[width=.48\textwidth]{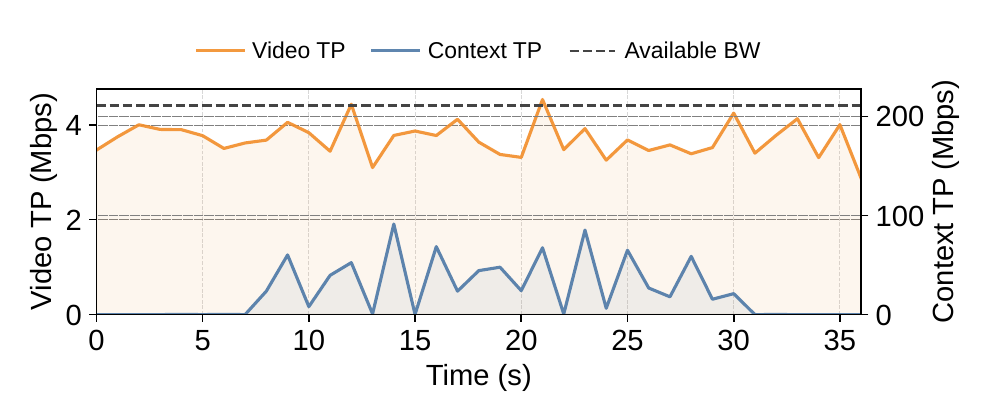}
    \caption{Microsoft Teams during concurrent video and an agent context transfer on Wi-Fi~6 link.}
    \label{fig:teams-timeline}
\end{figure}

\vspace{3pt}
\noindent
\textbf{Teams --- QUIC with application-level throttling.} Teams takes a fundamentally different transport choice: the agent context flow is carried over QUIC protocol~\cite{rfc9000} on UDP 443 rather than TCP. Figure~\ref{fig:teams-timeline} plots the per-second throughput of both flows across a 35-second session on a 211~Mbps Wi-Fi~6 link.

Although QUIC uses a modern queue-building CCA, such as Cubic or BBR, the throughput trace in Figure~\ref{fig:teams-timeline} does not resemble a CCA-driven transfer on an otherwise-empty 211~Mbps link.
Instead, we observe that an app-layer mechanism limits the rate with on/off burst patterns. Specifically, the QUIC flow alternates between sub-second bursts and near-zero pauses at roughly 1--2~s period.
This is inconsistent with QUIC CUBIC/BBR ramp-up behavior, which would sustain a high rate once probed. We conjecture that this operation is implemented by the app provider to protect RTP transmission.

\section{Rate Control Parameters}
\label{appendix:5-rate-control}

We detail the parameter choices in~\S\ref{subsec:5-app-guided} as follows. 


\vspace{3pt}
\noindent
\textbf{MI gain $\gamma_{\mathrm{mi}}$.}
\REV{
We use a queue-dependent MI gain to probe available bandwidth rapidly under low queuing and probe less aggressively as queuing increases.
Using the EWMA-smoothed queue estimate $\tilde{Q}_f$, we set
}
\begin{equation}
    \gamma_{\mathrm{mi}}
    =
    \gamma_{\mathrm{scale}}(\tilde{Q}_f)
    \cdot
    \max\!\left\{
        \rho,\;
        1-(1-\rho)\frac{\tilde{Q}_f}{Q_{\mathrm{floor}}}
    \right\}.
\end{equation}
\REV{
$\gamma_{\mathrm{scale}}(\tilde{Q}_f)$ is a queue-dependent probing scale, and $\rho$ is the minimum factor applied to that scale.
$Q_{\mathrm{floor}}$ is the smoothed queuing-delay threshold at which this factor reaches $\rho$.
The coefficient $1-\rho$ makes the factor decrease linearly from $1$ at $\tilde{Q}_f=0$ to $\rho$ at $\tilde{Q}_f=Q_{\mathrm{floor}}$.
For $\tilde{Q}_f \ge Q_{\mathrm{floor}}$, the factor remains at $\rho$ rather than decreasing further. We empirically set $\rho=0.25$ and $Q_{\mathrm{floor}}=0.9\cdot L$, and select the probing scale as}
\begin{equation}
    \gamma_{\mathrm{scale}}(\tilde{Q}_f)
    =
    \begin{cases}
        0.20, & \tilde{Q}_f < 0.1L,\\
        0.08, & \text{otherwise}.
    \end{cases}
\end{equation}
\REV{
The larger scale accelerates bandwidth exploration under low-queue conditions.
Setting $Q_{\mathrm{floor}}=0.9\cdot L$ places the threshold for applying the minimum factor below the nominal MI-to-AI boundary $L$.
This threshold does not end MI: the controller continues to increase the sending rate as long as $Q_f<L$.
}

\vspace{3pt}
\noindent
\textbf{AI gain $\gamma_{\mathrm{ai}}$.} We use $\gamma_{\mathrm{ai}} = 3$; a single-MSS step per RTT under-probes short-RTT paths where $L \gg \mathrm{RTT}_{\min}$.

\vspace{3pt}
\noindent
\textbf{AI threshold $\alpha$.}
We set $\alpha = 0.8$, so multiplicative decrease activates once the frame-level queuing is approaching its deadline. This leaves enough headroom for the proportional AI step to absorb transient overshoots without triggering Drain, while still reacting early enough to avoid deadline violations under sustained overload.

\vspace{3pt}
\noindent
\textbf{MD coefficient $\delta$.}
We set $\delta = 0.85$, a mild under-send that empties the queue within a few frames. A smaller $\delta$ drains faster but under-utilizes the link when the queue has only briefly exceeded the threshold.

\section{LoRA Training Details}
\label{appendix:lora-training}

\sysname{} uses LoRA-fine-tuned per-model length predictors that take a \{context, query\} pair and emit an integer estimate of the decoding length, following Zheng et al.~\cite{zheng2023response}. Table~\ref{tab:lora-config} summarizes the LoRA configuration: all three base LLMs receive adapters on every attention projection, with rank scaled to yield roughly 1\% trainable parameter overhead.

\begin{table}[t]
\centering
\footnotesize
\caption{Per-model LoRA configuration. Adapters are injected on \texttt{q\_proj}, \texttt{k\_proj}, \texttt{v\_proj}, and \texttt{o\_proj} of every decoder layer. We set $\alpha=2r$ and LoRA dropout $0.05$.}
\label{tab:lora-config}
\begin{tabular}{lcccr}
\toprule
Model & $r$ & $\alpha$ & Targets & Trainable \\
\midrule
Qwen2.5-3B (3.1B)  & 64  & 128 & q,k,v,o & 29.5M (0.95\%) \\
Qwen2.5-7B (7.6B)  & 128 & 256 & q,k,v,o & 80.7M (1.06\%) \\
Llama3.1-8B (8.0B) & 96 & 192 & q,k,v,o & 81.8M (1.02\%) \\
\bottomrule
\end{tabular}
\end{table}

\vspace{3pt}
\noindent\textbf{Training data.}
We pool roughly 4{,}500 long-context \{context, query\} pairs drawn from MuSiQue-Ans~\cite{trivedi2022musique}, QMSum~\cite{zhong2021qmsum}, and LongBench~\cite{bai2023longbench}, strictly disjoint from the 600 evaluation contexts to prevent leakage. For each base LLM we then re-label every pair with \emph{that model's own} greedy-decoded token count under the runtime configuration used at inference (KV cache pruned to ratio 0.3 via KVZip~\cite{kim2025kvzip}; \texttt{max\_new\_tokens}=128 for QA, 384 for summarization), yielding a per-model corpus of $\sim$3{,}700 $($prompt, integer-label$)$ pairs that we split 95/5 into train and validation partitions.

\vspace{3pt}
\noindent\textbf{Prompt and loss.}
Each training example uses the Stanford-Alpaca instruction template, with the ``\texttt{\#\#\# Instruction}'' field instructing the model to ``Estimate the number of tokens the response will contain \ldots\ reply with an integer only'' and the ``\texttt{\#\#\# Response}'' field containing the integer label. We mask all tokens preceding ``\texttt{\#\#\# Response:}'' via HuggingFace TRL's \texttt{DataCollatorForCompletionOnlyLM}, so the cross-entropy loss applies only to the 2--3 response tokens --- this is essential because without the mask, the signal from the integer label is diluted three orders of magnitude by $\sim\!3{,}000$ context tokens, and the LoRA learns only to emit \emph{some} integer rather than the correct one.

\vspace{3pt}
\noindent\textbf{Optimization.}
Training runs on a single NVIDIA GH200 GPU (96\,GB HBM) using \texttt{bf16} weights and gradients, gradient checkpointing (reentrant off), per-device batch size $1$, and gradient accumulation $8$ (effective batch 8). We use AdamW with learning rate $2\!\times\!10^{-4}$, a 10\% linear warm-up, and cosine decay to zero over 5 epochs ($2{,}200$ optimizer steps), which takes 2--3.5 hours depending on model size.

\begin{table}[t]
\centering
\small
\caption{Context size and decoding length distribution of our dataset (min--max, mean in parentheses).}
\label{tab:appendix-dataset-stats}
\setlength{\tabcolsep}{3pt}
\begin{tabular}{llccc}
\toprule
Dataset & Tokens & Qwen2.5-3B & Qwen2.5-7B & Llama3.1-8B \\
\midrule
\multirow{2}{*}{QA}
  & Context & \multicolumn{3}{c}{1k--4k (2{,}532)} \\
  \cmidrule(lr){2-5}
  & Decoding & 1--126 (44)   & 1--113 (34)   & 1--56 (7)     \\
\midrule
\multirow{2}{*}{Summ.}
  & Context & \multicolumn{3}{c}{8k--16k (11{,}985)} \\
  \cmidrule(lr){2-5}
  & Decoding & 70--374 (170) & 100--381 (210) & 31--380 (129) \\
\bottomrule
\end{tabular}
\end{table}

\section{Dataset Statistics}
\label{appendix:dataset-stats}

Table~\ref{tab:appendix-dataset-stats} summarizes the full per-dataset context and per-model decoding-length distributions used in \S\ref{subsec:8-setup}.

\section{\REV{Multi-Agent Flow Coordinator Microbenchmark}}
\label{appendix:multi-agent-microbenchmark}

\begin{figure}[t]
  \centering
  \begin{subfigure}[t]{0.48\textwidth}
    \centering
    \includegraphics[width=\textwidth]{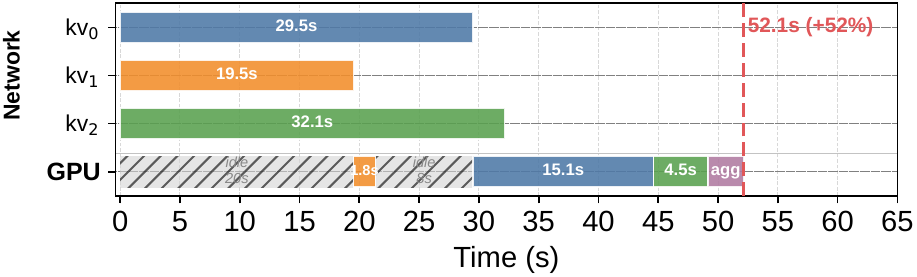}
    \vspace{-15pt}
    \caption{Round robin.}
    \label{fig:8-eval-microbenchmark-multi-agent_a}
  \end{subfigure}%
  \hfill
  \begin{subfigure}[t]{0.48\textwidth}
    \centering
    \includegraphics[width=\textwidth]{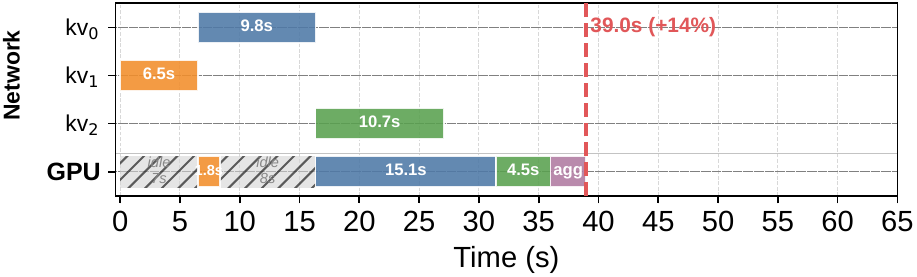}
    \vspace{-15pt}
    \caption{Shortest network job first.}
    \label{fig:8-eval-microbenchmark-multi-agent_b}
  \end{subfigure}
  \hfill
  \begin{subfigure}[t]{0.48\textwidth}
    \centering
    \includegraphics[width=\textwidth]{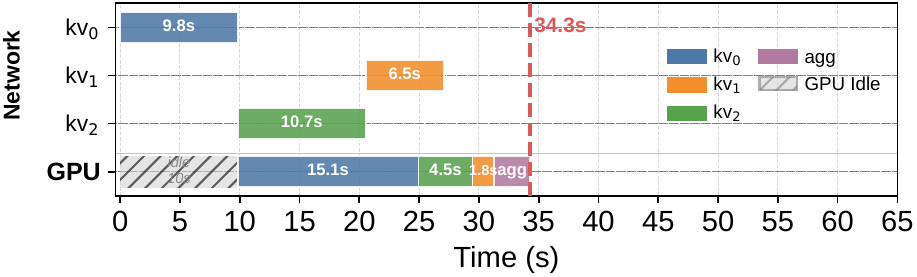}
    \vspace{-15pt}
    \caption{\sysname{}.}
    \label{fig:8-eval-microbenchmark-multi-agent_c}
  \end{subfigure}
  \vspace{-12pt}
  \caption{Multi-agent flow coordination timeline.}
  \label{fig:8-eval-microbenchmark-multi-agent}
  \vspace{-5pt}
\end{figure}

\REV{
Figure~\ref{fig:8-eval-microbenchmark-multi-agent} compares the scheduling timelines of three multi-agent flow coordination policies (round robin, shortest network job first, and \sysname{}) on the same three-context summarization workload.
Round robin delays the completion of all three context transfers, leaving the GPU idle for extended periods and increasing the agent response time to 52.1\,s.
Shortest network job first prioritizes the smallest context transfer ($kv_1$), allowing the GPU to begin decoding sooner.
However, decoding for $kv_1$ finishes before the next context transfer ($kv_0$) completes, resulting in $\approx$8\,s GPU idle time. 
By jointly considering flow sizes and decoding latencies, \sysname{} reduces the response time by 17.8\,s compared with round robin.
}

\begin{figure}[t]
  \centering
  \begin{minipage}[t]{0.23\textwidth}
    \centering
    \includegraphics[width=\textwidth]{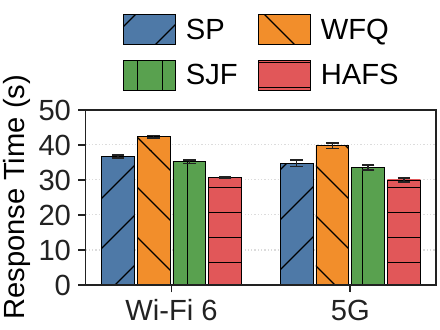}
    \caption{\REV{Comparison with various multi-agent flow scheduling policies.}}
    \label{fig:appendix:scheduler_compare}
  \end{minipage}
  \hfill
  \begin{minipage}[t]{0.23\textwidth}
    \centering
    \includegraphics[width=\textwidth]{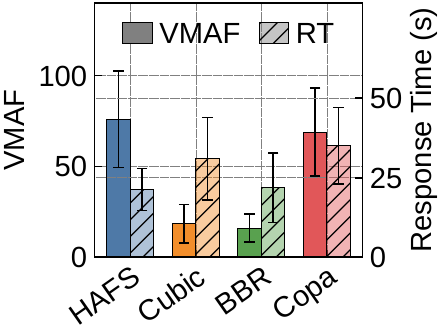}
    \caption{\REV{Comparison of application performance across TCP CCAs.}}
    \label{fig:appendix:cca_substitution}
  \end{minipage}
\end{figure}

\REV{
Figure~\ref{fig:appendix:scheduler_compare} further compares \sysname{}'s agent response times with those under three scheduling policies: (i) Strict Priority (SP), which assigns static user-level priorities and uses round-robin scheduling across agent flows belonging to the same user, (ii) Weighted Fair Queuing (WFQ), which allocates bandwidth among agent flows sharing the same bottleneck link in proportion to their flow sizes, and (iii) Shortest Network Job First (SJF).
Contention among agent flows decreases from WFQ to SP to SJF, reducing GPU idle time (and thus the agent response time). 
\sysname{} consistently achieves the lowest response time across different network traces (13\% lower than that of SJF), highlighting the benefit of jointly considering network and compute latency.
}

\section{\REV{Application Performance Comparison with Various CCAs}}
\label{appendix:ablation-cca}

\REV{
Figure~\ref{fig:appendix:cca_substitution} compares video quality and agent response time under \sysname{}, CUBIC, BBR, and Copa during concurrent video streaming and KV-cache transfers.
For the TCP baselines, we estimate the aggregate TCP delivery rate from per-connection \texttt{tcpi\_delivery\_rate} values obtained via \texttt{TCP\_INFO} and set GCC's minimum bitrate to $b_{\mathrm{req}}$ when the estimate is larger (as in~\S\ref{subsec:5-app-guided}), while leaving TCP rate control to the selected CCA.
CUBIC and BBR yield substantially lower VMAF than \sysname{} because the GCC bitrate floor alone does not prevent queue buildup during KV-cache transfers from delaying video frames beyond their deadlines.
Copa better preserves video quality but incurs the longest response time.
By adapting agent rates to frame-level queuing and video deadlines, \sysname{} limits queue buildup while maintaining high throughput, achieving highest VMAF (average 3.7$\times$ compared to BBR) and lowest response time (average 39.3\% lower compared to Copa).
}

\section{Fairness of \sysname{}}
\label{appendix:fairness}

\begin{figure}[t]
  \centering
  \begin{subfigure}[t]{0.23\textwidth}
    \centering
    \includegraphics[width=\textwidth]{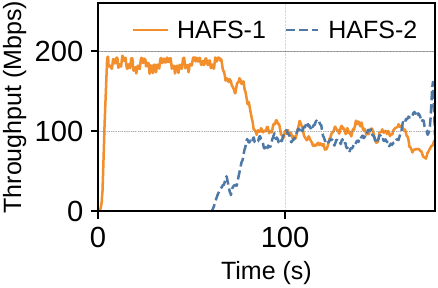}
    \vspace{-20pt}
    \caption{\sysname{} vs.\ \sysname{}.}
    \label{fig:7-fairness-hafs}
  \end{subfigure}
  \hfill
  \begin{subfigure}[t]{0.23\textwidth}
    \centering
    \includegraphics[width=\textwidth]{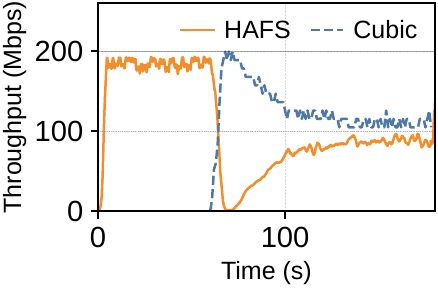}
    \vspace{-20pt}
    \caption{\sysname{} vs.\ Cubic.}
    \label{fig:7-fairness-cubic}
  \end{subfigure}
  \vspace{-5pt}
  \caption{Fairness behavior of \sysname{} over 200~Mbps fixed bandwidth.}
  \label{fig:hafs-fairness}
\end{figure}

As shown in Figure~\ref{fig:hafs-fairness}, when two \sysname{} flows share a bottleneck with the same frame-level deadline, their rate controllers converge to the fair-share rates, as both flows react symmetrically to the shared per-frame queuing signal.
However, when a \sysname{} flow competes with a queue-building CCA, Cubic, the shared queue inflates and the per-frame queuing signal rises.
\sysname{} deliberately yields rate in this regime---staying just within the RTP deadline rather than contending for the queue---so that the latency-critical video flow is protected even at the cost of throughput for the background flow.

\end{document}